\documentclass{article}

    \usepackage[final]{neurips_2022}

\usepackage{makecell}
\usepackage[utf8]{inputenc} % allow utf-8 input
\usepackage[T1]{fontenc}    % use 8-bit T1 fonts
\usepackage{hyperref}       % hyperlinks
\usepackage{url}            % simple URL typesetting
\usepackage{booktabs}       % professional-quality tables
\usepackage{amsfonts}       % blackboard math symbols
\usepackage{nicefrac}       % compact symbols for 1/2, etc.
\usepackage{microtype}      % microtypography
\usepackage{xcolor}         % colors
\usepackage{graphicx} 
\usepackage{amsmath}
\usepackage{algorithm}
\usepackage{algpseudocode}
\usepackage{mathtools}
\usepackage{mathrsfs}
\usepackage{caption}
\usepackage{subcaption}

\newcommand{\comment}[1]{}

\newcommand{\modelname}{Shap$\cdot$E}
\newcommand{\pointe}{Point$\cdot$E}
\newcommand{\shortcite}[1]{[\citenum{#1}]}
\newcommand{\namecite}[1]{\citeauthor{#1} [\citenum{#1}]}

\title{\modelname{}: Generating Conditional 3D Implicit Functions}

\makeatletter
\newcommand{\printfnsymbol}[1]{%
  \textsuperscript{\@fnsymbol{#1}}%
}
\makeatother

\author{%
  Heewoo Jun \thanks{Equal contribution} \\
  \texttt{heewoo@openai.com} \\
  \And
  Alex Nichol \printfnsymbol{1} \\
  \texttt{alex@openai.com} \\
}

\begin{document}

\maketitle

\begin{abstract}
    We present \modelname{}, a conditional generative model for 3D assets. Unlike recent work on 3D generative models which produce a single output representation, \modelname{} directly generates the parameters of implicit functions that can be rendered as both textured meshes and neural radiance fields. We train \modelname{} in two stages: first, we train an encoder that deterministically maps 3D assets into the parameters of an implicit function; second, we train a conditional diffusion model on outputs of the encoder. When trained on a large dataset of paired 3D and text data, our resulting models are capable of generating complex and diverse 3D assets in a matter of seconds. When compared to \pointe{}, an explicit generative model over point clouds, \modelname{} converges faster and reaches comparable or better sample quality despite modeling a higher-dimensional, multi-representation output space. We release model weights, inference code, and samples at \url{https://github.com/openai/shap-e}.
\end{abstract}

\section{Introduction}

\begin{figure}[hbtp]
    \centering
    \setlength{\tabcolsep}{2.0pt}
    \begin{tabular}{cccc}
        \includegraphics[width=0.25\textwidth]{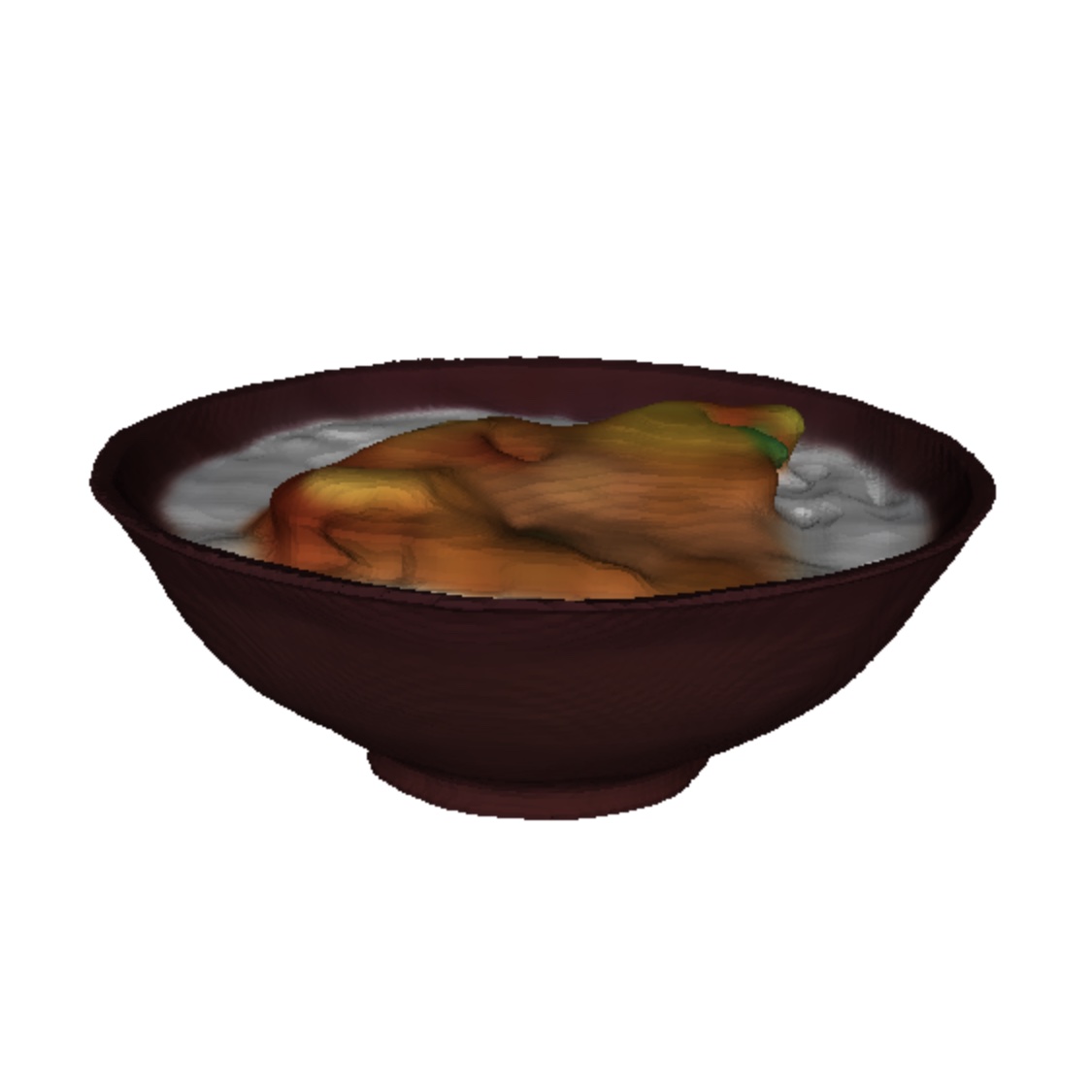} &
        \includegraphics[width=0.25\textwidth]{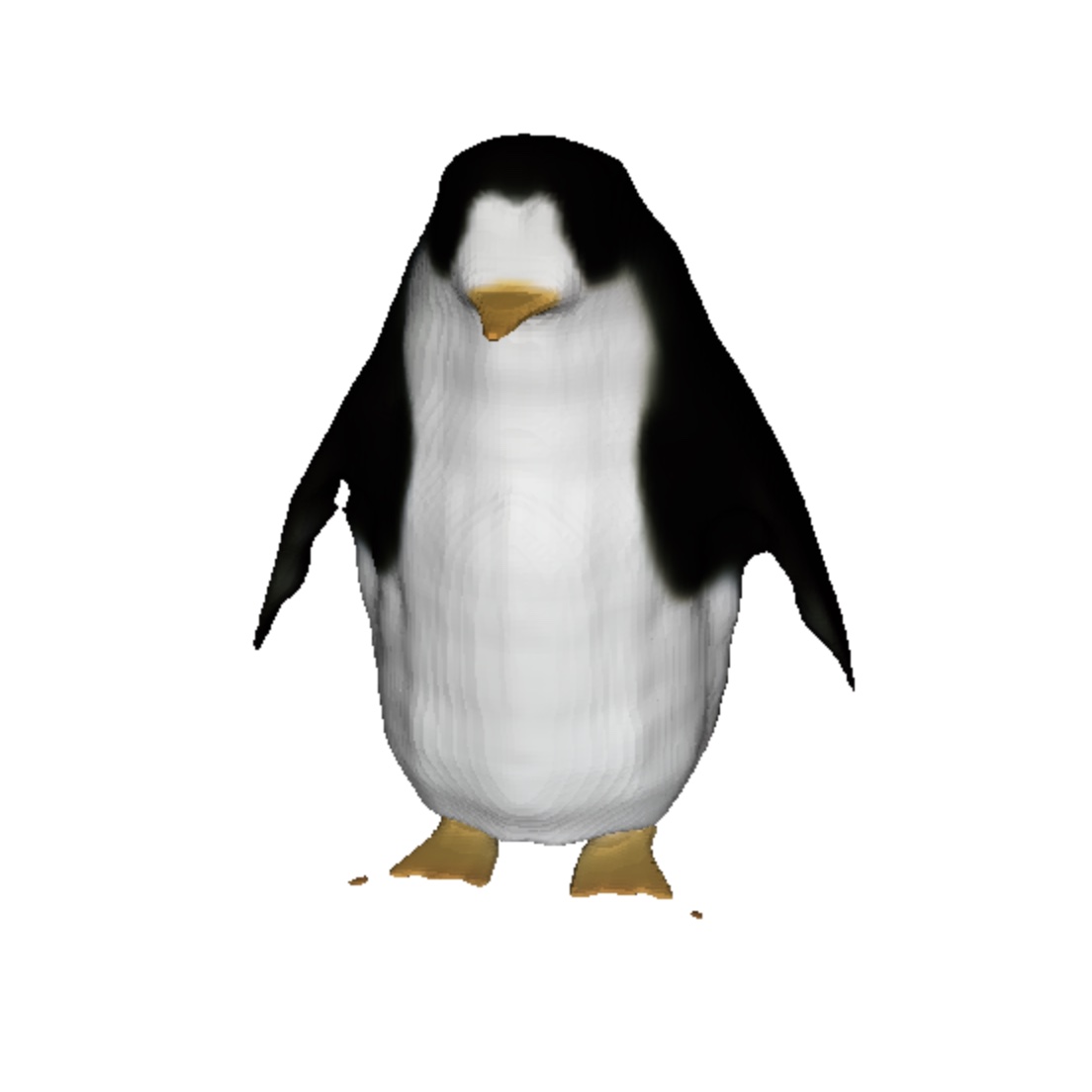} &
        \includegraphics[width=0.25\textwidth]{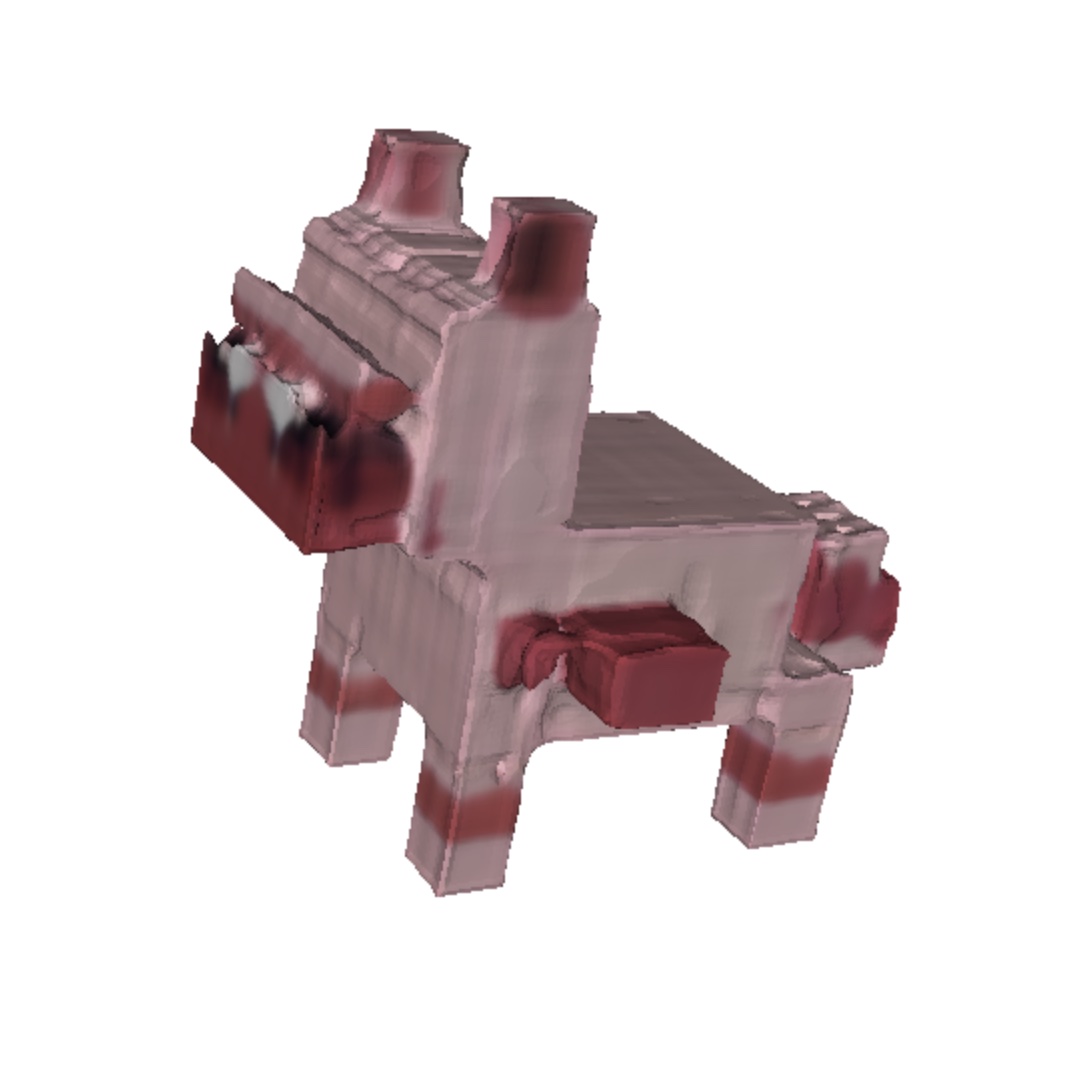} &
        \includegraphics[width=0.25\textwidth]{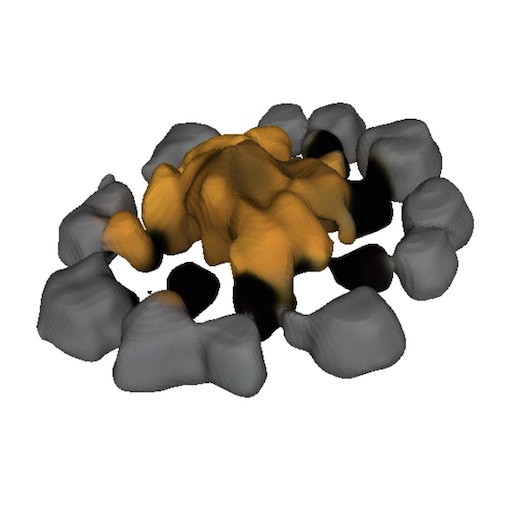} \\

        \scriptsize \makecell{``a bowl of food''} &
        \scriptsize \makecell{``a penguin''} &
        \scriptsize \makecell{``a voxelized dog''} &
        \scriptsize \makecell{``a campfire''} \\
        % \rule{0pt}{0.15pt} \\

        \includegraphics[width=0.25\textwidth]{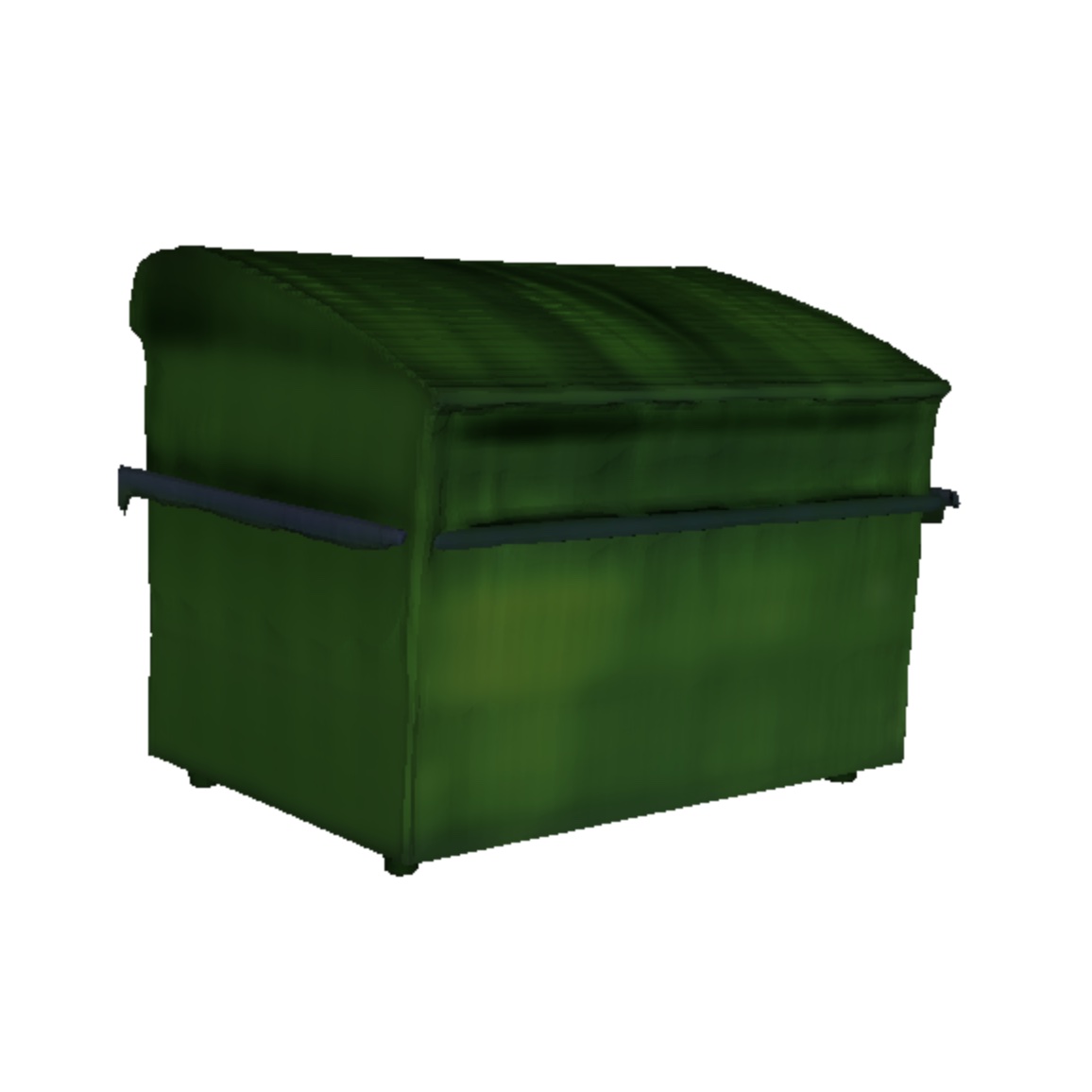} &
        \includegraphics[width=0.25\textwidth]{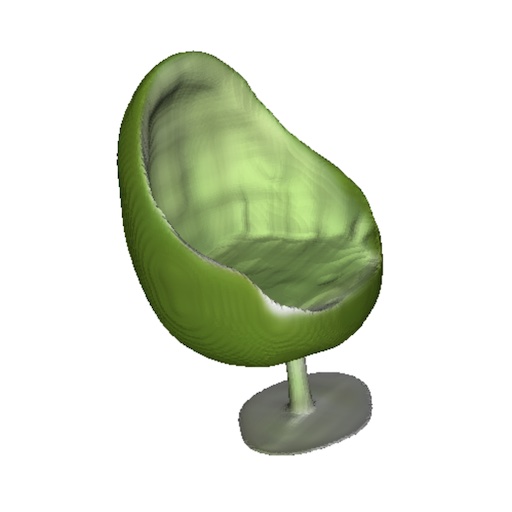} &
        \includegraphics[width=0.25\textwidth]{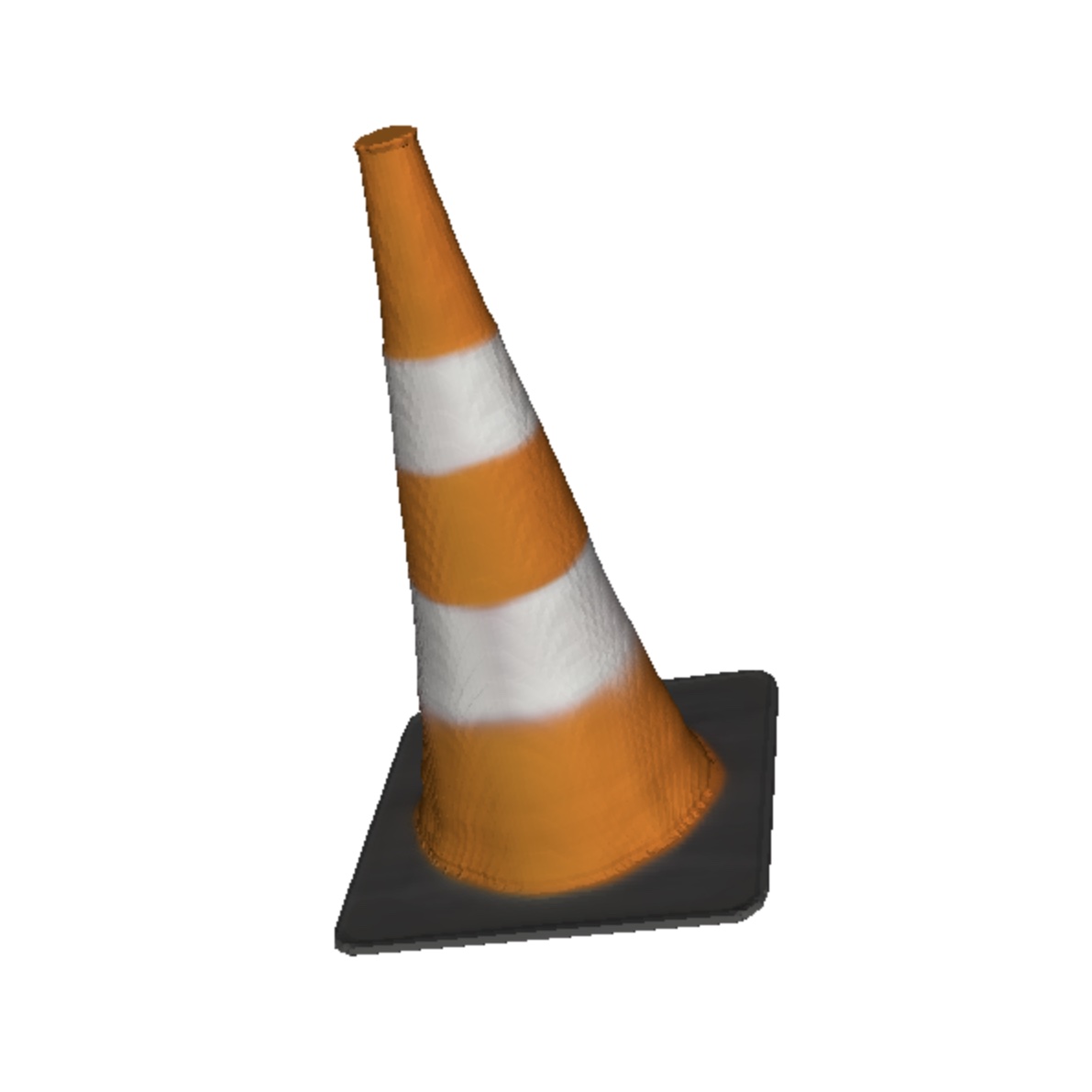} &
        \includegraphics[width=0.25\textwidth]{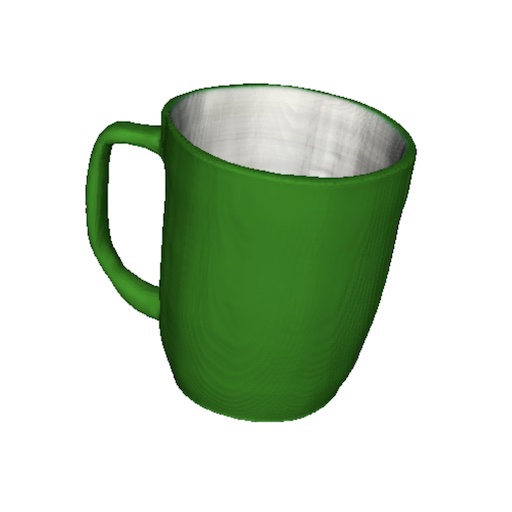} \\

        \scriptsize \makecell{``a dumpster''} &
        \scriptsize \makecell{``a chair that looks like \\ an avocado''} &
        \scriptsize \makecell{``a traffic cone''} &
        \scriptsize \makecell{``a green coffee mug''} \\
        % \rule{0pt}{0.15pt} \\

        \includegraphics[width=0.25\textwidth]{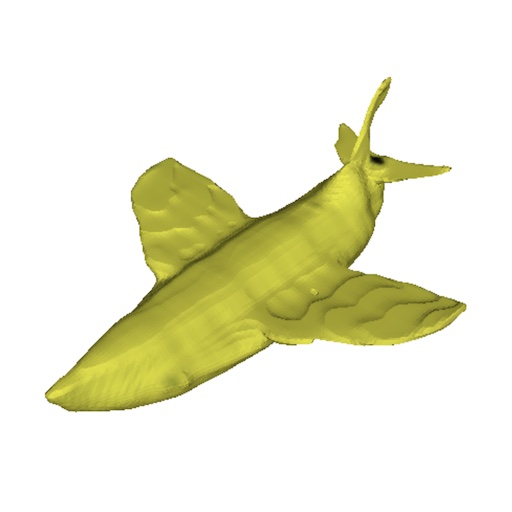} &
        \includegraphics[width=0.25\textwidth]{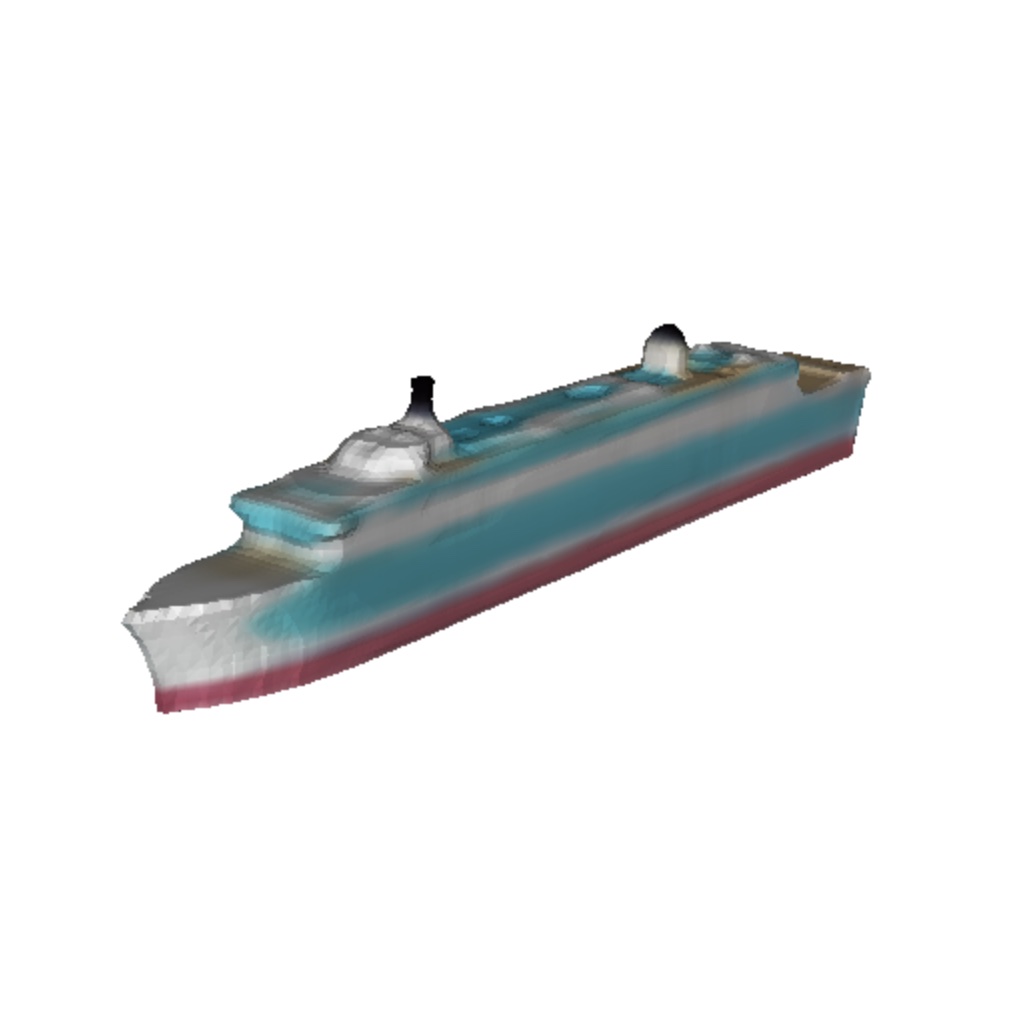} &
        \includegraphics[width=0.25\textwidth]{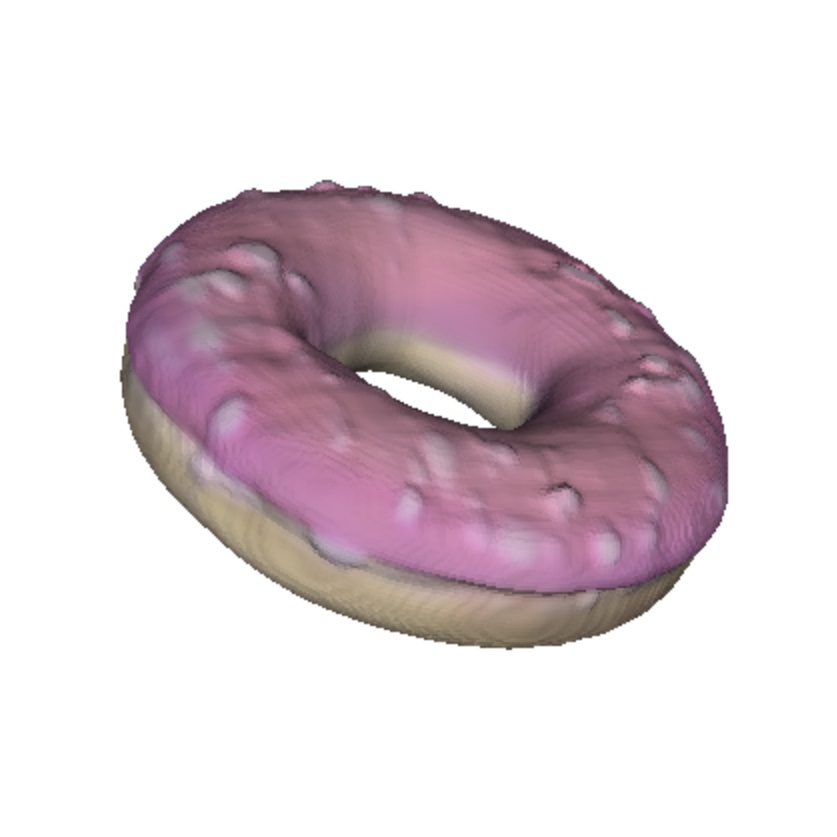} &
        \includegraphics[width=0.25\textwidth]{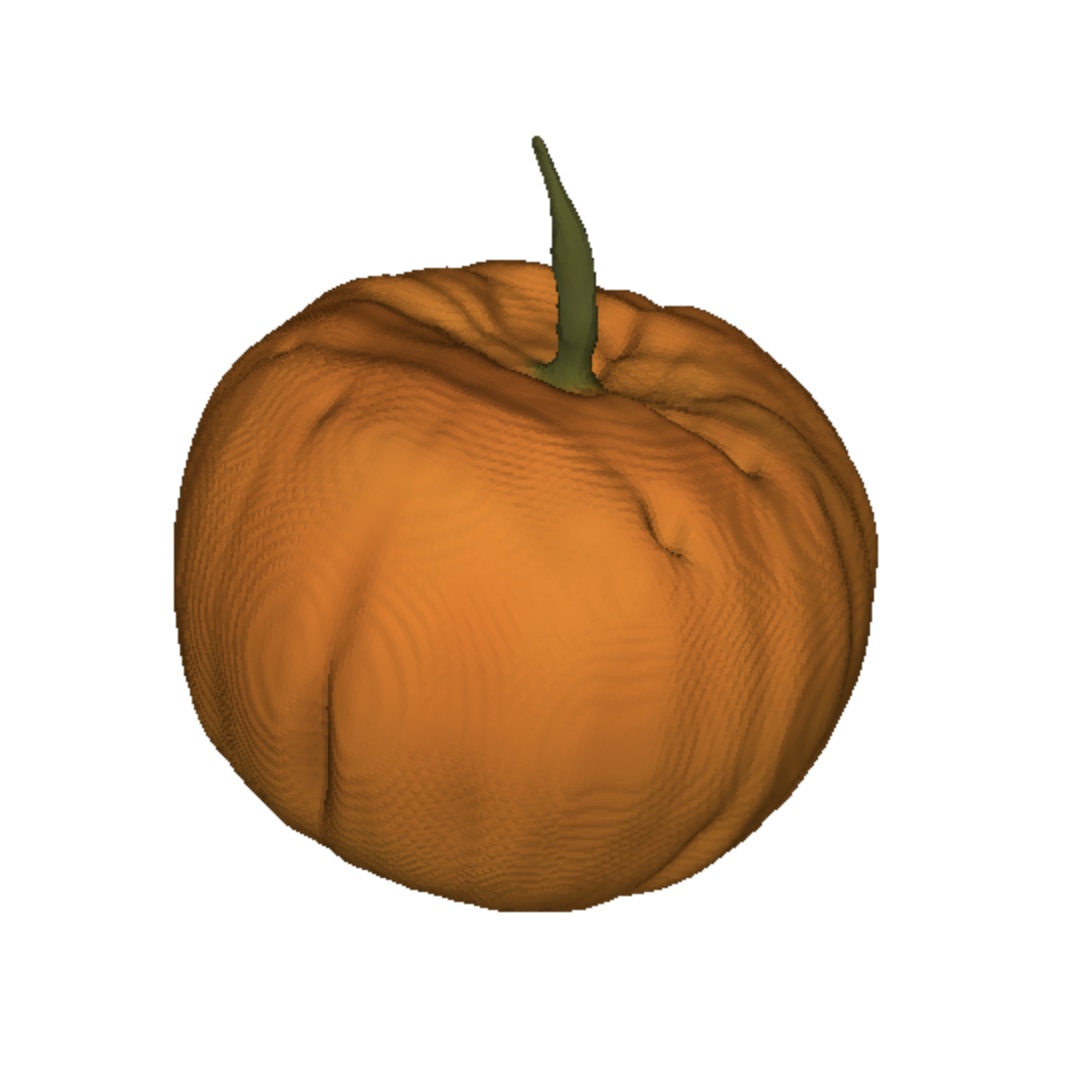} \\

        \scriptsize \makecell{``an airplane that looks \\ like a banana''} &
        \scriptsize \makecell{``a cruise ship''} &
        \scriptsize \makecell{``a donut with pink icing''} &
        \scriptsize \makecell{``a pumpkin''} \\
        % \rule{0pt}{0.15pt} \\
        \includegraphics[width=0.25\textwidth]{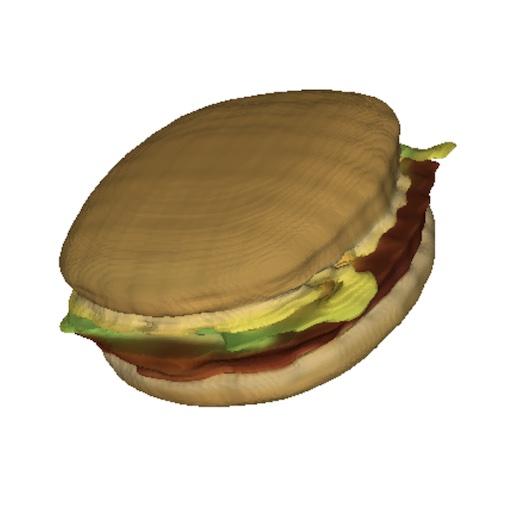} &
        \includegraphics[width=0.25\textwidth]{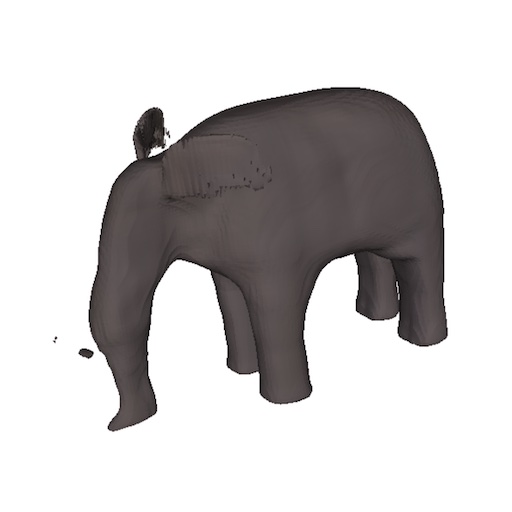} &
        \includegraphics[width=0.25\textwidth]{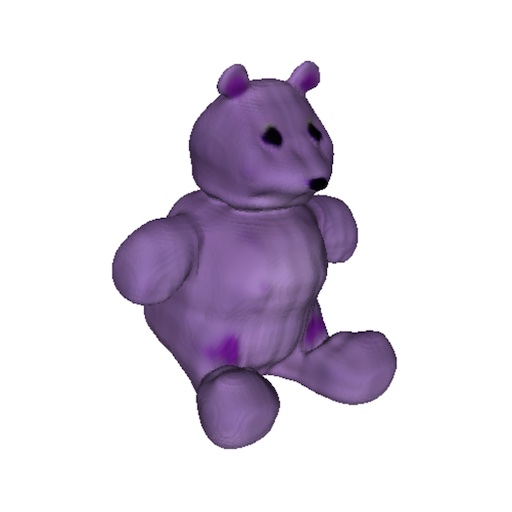} &
        \includegraphics[width=0.25\textwidth]{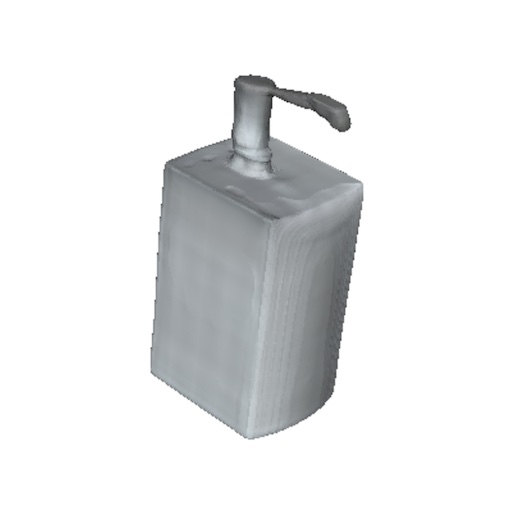} \\

        \scriptsize \makecell{``a cheeseburger''} &
        \scriptsize \makecell{``an elephant''} &
        \scriptsize \makecell{``a light purple teddy bear''} &
        \scriptsize \makecell{``a soap dispenser''} \\

        % \rule{0pt}{0.15pt} \\
        \includegraphics[width=0.25\textwidth]{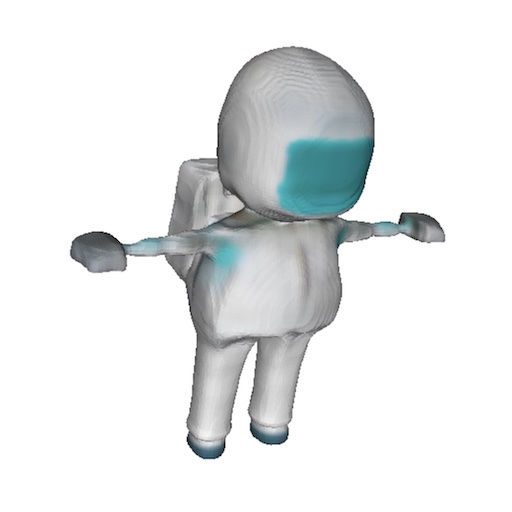} &
        \includegraphics[width=0.25\textwidth]{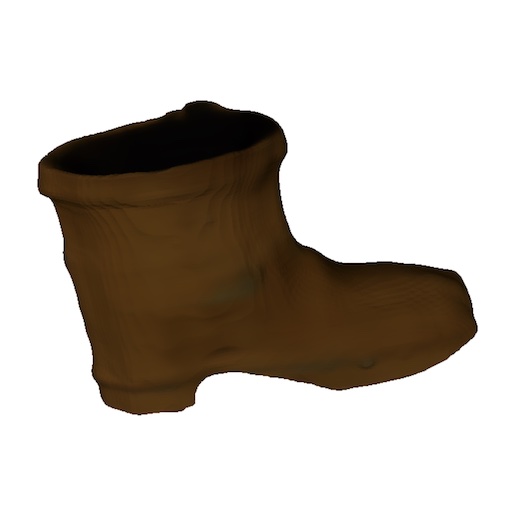} &
        \includegraphics[width=0.25\textwidth]{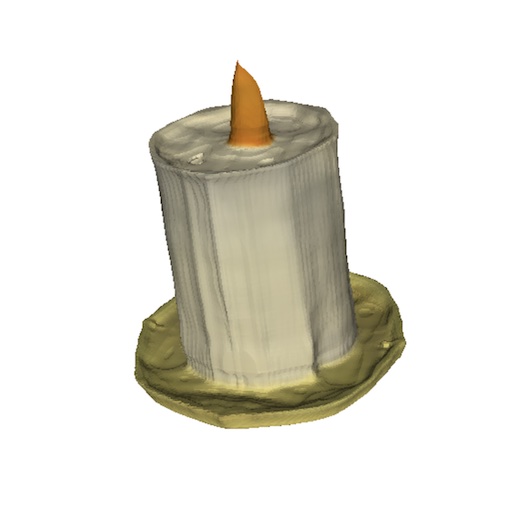} &
        \includegraphics[width=0.25\textwidth]{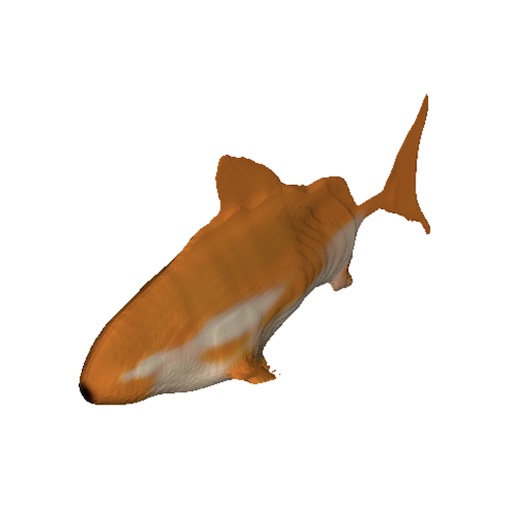} \\

        \scriptsize \makecell{``an astronaut''} &
        \scriptsize \makecell{``a brown boot''} &
        \scriptsize \makecell{``a lit candle''} &
        \scriptsize \makecell{``a goldfish''} \\
    \end{tabular}

    \caption{\label{fig:banner_images} Selected text-conditional meshes generated by \modelname{}. Each sample takes roughly 13 seconds to generate on a single NVIDIA V100 GPU, and does not require a separate text-to-image model.}
    \vskip -0.1in
\end{figure}

With the recent explosion of generative image models \shortcite{dalle,glide,unclip,makeascene,latentdiffusion,imagen,parti,engievilg2}, there has been increasing interest in training similar generative models for other modalities such as audio \shortcite{musenet,jukebox,audiolm,audiogen,musiclm,noise2music}, video \shortcite{videodm,makeavideo,imagenvideo}, and 3D assets \shortcite{dreamfields,clipmesh,dreamfusion,pointe}. Most of these modalities lend themselves to natural, fixed-size tensor representations that can be directly generated, such as grids of pixels for images or arrays of samples for audio. However, it is less clear how to represent 3D assets in a way that is efficient to generate and easy to use in downstream applications.

Recently, implicit neural representations (INRs) have become popular for encoding 3D assets. To represent a 3D asset, INRs typically map 3D coordinates to location-specific information such as density and color. In general, INRs can be thought of as resolution independent, since they can be queried at arbitrary input points rather than encoding information in a fixed grid or sequence. Since they are end-to-end differentiable, INRs also enable various downstream applications such as style transfer \shortcite{artisticnerf} and differentiable shape editing \shortcite{semanticnerf}. In this work, we focus on two types of INRs for 3D representation:

\begin{itemize}
    \item \textbf{A Neural Radiance Field (NeRF)} \shortcite{nerf} is an INR which represents a 3D scene as a function mapping coordinates and viewing directions to densities and RGB colors. A NeRF can be rendered from arbitrary views by querying densities and colors along camera rays, and trained to match ground-truth renderings of a 3D scene.
    \item \textbf{DMTet} \shortcite{deepmarchingtets} and its extension \textbf{GET3D} \shortcite{get3d} represent a textured 3D mesh as a function mapping coordinates to colors, signed distances, and vertex offsets. This INR can be used to construct 3D triangle meshes in a differentiable manner, and the resulting meshes can be rendered efficiently using differentiable rasterization libraries \shortcite{nvdiffrast}.
\end{itemize}

Although INRs are flexible and expressive, the process of acquiring them for each sample in a dataset can be costly. Additionally, each INR may have many numerical parameters, potentially posing challenges when training downstream generative models.
Some works approach these issues by using auto-encoders with an implicit decoder to obtain smaller latent representations that can be directly modeled with existing generative techniques \shortcite{deepsdf,towardsimplicit,nerfvae}.
\namecite{functa} present an alternative approach, where they use meta-learning to create a dataset of INRs that share most of their parameters, and then train diffusion models \shortcite{dickstein,scorematching,ddpm} or normalizing flows \shortcite{flows,splineflows} on the free parameters of these INRs. \namecite{inr-encoder} further suggest that gradient-based meta-learning might not be necessary at all, instead directly training a Transformer \shortcite{transformer} encoder to produce NeRF parameters conditioned on multiple views of a 3D object.

We combine and scale up several of the above approaches to arrive at \modelname{}, a conditional generative model for diverse and complex 3D implicit representations. First, we scale up the approach of \namecite{inr-encoder} by training a Transformer-based encoder to produce INR parameters for 3D assets. Next, similar to \namecite{functa}, we train a diffusion model on outputs from the encoder. Unlike previous approaches, we produce INRs which represent both NeRFs and meshes simultaneously, allowing them to be rendered in multiple ways or imported into downstream 3D applications.

When trained on a dataset of several million 3D assets, our models are capable of producing diverse, recognizable samples conditioned on text prompts (Figure \ref{fig:banner_images}). Compared to \pointe{} \shortcite{pointe}, a recently proposed explicit 3D generative model, our models converge faster and obtain comparable or superior results while sharing the same model architecture, datasets, and conditioning mechanisms. 

Surprisingly, we find that \modelname{} and \pointe{} tend to share success and failure cases when conditioned on images, suggesting that very different choices of output representation can still lead to similar model behavior. However, we also observe some qualitative differences between the two models, especially when directly conditioning on text captions. Like \pointe{}, the sample quality of our models still falls short of optimization-based approaches for text-conditional 3D generation. However, it is orders of magnitude faster at inference time than these approaches, allowing for a potentially favorable trade-off.

We release our models, inference code, and samples at \url{https://github.com/openai/shap-e}.

\section{Background}

\subsection{Neural Radiance Fields (NeRF)}

\namecite{nerf} introduce NeRF, a method for representing a 3D scene as an implicit function defined as
$$F_{\Theta} : (\mathbf{x}, \mathbf{d}) \mapsto (\mathbf{c}, \sigma)$$
where $\mathbf{x}$ is a 3D spatial coordinate, $\mathbf{d}$ is a 3D viewing direction, $\mathbf{c}$ is an RGB color, and $\sigma$ is a non-negative density value. For convenience, we split $F_{\Theta}$ into separate functions, $\sigma(\mathbf{x})$ and $\mathbf{c}(\mathbf{x}, \mathbf{d})$.

To render a novel view of a scene, we treat the viewport as a grid of rays and render each ray by querying $F_{\Theta}$ at points along the ray. More precisely, each pixel of the viewport is assigned a ray $\mathbf{r}(t) = \mathbf{o} + t\mathbf{d}$ which extends from the camera origin $\mathbf{o}$ along a direction $\mathbf{d}$. The ray can then be rendered to an RGB color by approximating the integral
$$\hat{C}(\mathbf{r})=\int_{0}^{\infty} T(t)\sigma(\mathbf{r}(t))\mathbf{c}(\mathbf{r}(t),\mathbf{d}) dt \text{ , where } T(t) = \text{exp}\left(-\int_{0}^t \sigma(\mathbf{r}(s))ds\right)$$

\namecite{nerf} use quadrature to approximate this integral. In particular, they define a sequence of increasing values $t_i, i \in [1, N]$ and corresponding $\delta_i = t_{i+1} - t_i$. The integral is then approximated via a discrete sum
$$\hat{C}(\mathbf{r}) = \sum_{i=1}^{N} T_i(1-\text{exp}(-\sigma(\mathbf{r}(t_i)) \delta_i)) \mathbf{c}(\mathbf{r}(t_i), \mathbf{d})\text{, where } T_i = \text{exp}\left(-\sum_{j=1}^{i-1} \sigma(\mathbf{r}(t_j)) \delta_j \right)$$

One remaining question is how to select the sequence of $t_0,...t_N$ to achieve an accurate estimate. This can be especially important for thin features, where a coarse sampling of points along the ray may completely miss a detail of the object. To address this problem, \namecite{nerf} suggest a two-stage rendering procedure. In the first stage, timesteps $t_i$ are sampled along uniform intervals of a ray, giving a coarse estimate of the predicted color $\hat{C}_c$. In computing this integral, they also compute weights proportional to the influence of each point along the ray:
$$w_i \sim T_i(1-\text{exp}(-\sigma(\mathbf{r}(t_i)) \delta_i))$$

To sample timesteps for the fine rendering stage, \namecite{nerf} use $w_i$ to define a piecewise-constant PDF along the ray. This allows a new set of $t_i$ to be sampled around points of high density in the scene. While \namecite{nerf} use two separate NeRF models for the coarse and fine rendering stages, we instead share the parameters between the two stages but use separate output heads for the coarse and fine densities and colors.

For notational convenience in later sections, we additionally define the transmittance of a ray as follows. Intuitively, this is the complement of the opacity or alpha value of a ray:
$$\hat{T}(\mathbf{r}) = 1 - \text{exp}\left(-\sum_{i=1}^{N} \sigma(\mathbf{r}(t_i)) \delta_i\right)$$

\subsection{Signed Distance Functions and Texture Fields (STF)}

Throughout this paper, we use the abbreviation \textit{STF} to refer to an implicit function which produces both signed distances and texture colors. This section gives some background on how these implicit functions can be used to construct meshes and produce renderings.

Signed distance functions (SDFs) are a classic way to represent a 3D shape as a scalar field. In particular, an SDF $f$ maps a coordinate $\mathbf{x}$ to a scalar $f(\mathbf{x}) = d$, such that $|d|$ is the distance of $\mathbf{x}$ to the nearest point on the surface of the shape, and $d < 0$ if the point is outside of the shape. As a result of this definition, the level set $f(\mathbf{x})=0$ defines the boundary of the shape, and $\text{sign}(d)$ determines normal orientation along the boundary. Methods such as marching cubes \shortcite{marchingcubes} or marching tetrahedra \shortcite{marchingtets} can be used to construct meshes from this level set.

\namecite{deepmarchingtets} present DMTet, a generative model over 3D shapes that leverages SDFs. DMTet produces SDF values $s_i$ and displacements $\Delta v_i$ for each vertex $v_i$ in a dense spatial grid. The SDF values are fed through a differentiable marching tetrahedra implementation to produce an initial mesh, and then the resulting vertices are offset using the additional vector $\Delta v_i$. They also employ a subdivision procedure to efficiently obtain more detailed meshes, but we do not consider this in our work for the sake of simplicity.

\namecite{get3d} propose GET3D, which augments DMTet with additional texture information. In particular, they train a separate model to predict RGB colors $\mathbf{c}$ for each surface point $\mathbf{p}$. This implicit model can be queried at surface points during rendering, or offline to construct an explicit texture. GET3D uses a differentiable rasterization library \shortcite{nvdiffrast} to produce rendered images for generated meshes. This provides an avenue to train the implicit function end-to-end with only image-space gradients.

\subsection{Diffusion Models}

Our work leverages denoising diffusion \shortcite{dickstein,scorematching,ddpm} to model a high-dimensional continuous distribution. We employ the Gaussian diffusion setup of \namecite{ddpm}, which defines a diffusion process that begins at a data sample $x_0$ and gradually applies Gaussian noise to arrive at increasingly noisy samples $x_1, x_2, ..., x_T$. Typically, the noising process is set up such that $x_T$ is almost indistinguishable from Gaussian noise. In practice, we never run the noising process sequentially, but instead ``jump'' directly to a noised version of a sample according to
$$x_t = \sqrt{\bar{\alpha}_t}x_0 + \sqrt{1-\bar{\alpha}_t} \epsilon$$

where $\epsilon \sim \mathcal{N}(0, \mathbf{I})$ is random noise, and $\bar{\alpha}_t$ is a monotonically decreasing noise schedule such that $\bar{\alpha}_0 = 1$. \namecite{ddpm} train a model $\epsilon_{\theta}(x_t, t)$ on a data distribution $q(x_0)$ by minimizing the objective:
$$L_{\text{simple}} = E_{x_0 \sim q(x_0), \epsilon \sim \mathcal{N}(0, \mathbf{I}), t \sim U[1, T]} ||\epsilon - \epsilon_{\theta}(x_t, t)||_2^2$$

However, \namecite{ddpm} also note an alternative but equivalent parameterization of the diffusion model, which we use in our work. In particular, we parameterize our model as $x_{\theta}(x_t, t)$ and train it to directly predict the denoised sample $x_0$ by minimizing
$$L_{x_0} = E_{x_0 \sim q(x_0), \epsilon \sim \mathcal{N}(0, \mathbf{I}), t \sim U[1, T]} ||x_{\theta}(x_t, t) - x_0||_2^2$$

To sample from a diffusion model, one starts at a random noise sample $x_T$ and gradually denoises it into samples $x_{T-1},...,x_0$ to obtain a sample $x_0$ from the approximated data distribution. While early work on these models focused on stochastic sampling processes \shortcite{dickstein,scorematching,ddpm}, other works propose alternative sampling methods which often draw on the relationship between diffusion models and ordinary differential equations \shortcite{ddim,sde}. In our work, we employ the Heun sampler proposed by \namecite{edm}, as we found it to produce high-quality samples with reasonable latency.

For conditional diffusion models, it is possible to improve sample quality at the cost of diversity using a guidance technique. \namecite{sotapaper} first showed this effect using image-space gradients from a noise-aware classifier, and \namecite{uncond} later proposed classifier-free guidance to remove the need for a separate classifier. To utilize classifier-free guidance, we train our diffusion model to condition on some information $y$ (e.g. a conditioning image or textual description), but randomly drop this signal during training to enable the model to make unconditional predictions. During sampling, we then adjust our model prediction as follows:

$$\hat{x}_{\theta}(x_t, t | y) = x_{\theta}(x_t, t) + s \cdot (x_{\theta}(x_t, t | y) - x_{\theta}(x_t, t))$$

where $s$ is a guidance scale. When $s = 0$ or $s = 1$, this is equivalent to regular unconditional or conditional sampling, respectively. Setting $s > 1$ typically produces more coherent but less diverse samples. We employ this technique for all of our models, finding (as expected) that guidance is necessary to obtain the best results.

\subsection{Latent Diffusion}

While diffusion can be applied to any distribution of vectors, it is often applied directly to signals such as pixels of images. However, it is also possible to use diffusion to generate samples in a continuous latent space.

\namecite{latentdiffusion} propose Latent Diffusion Models (LDMs) as a two-stage generation technique for images. Under the LDM framework, they first train an encoder to produce latents $z = E(x)$ and a decoder to produce reconstructions $\Tilde{x} = D(z)$. The encoder and decoder are trained in tandem to minimize a perceptual loss between $\Tilde{x}$ and $x$, as well as a patchwise discriminator loss on $\Tilde{x}$. After these models are trained, a second diffusion model is trained directly on encoded dataset samples. In particular, each dataset example $x_i$ is encoded into a latent $z_i$, and then $z_i$ is used as a training example for the diffusion model. To generate new samples, the diffusion model first generates a latent sample $z$, and then $D(z)$ yields an image. In the original LDM setup, the latents $z$ are lower-dimensional than the original images, and \namecite{latentdiffusion} propose to either regularize $z$ towards a normal distribution using a KL penalty, or to apply a vector quantization layer \shortcite{vqvae} to prevent $z$ from being difficult to model.

Our work leverages the above approach, but makes several simplifications. First, we do not use a perceptual loss or a GAN-based objective for our reconstructions, but rather a simple $L_1$ or $L_2$ reconstruction loss. Additionally, instead of using KL regularization or vector quantization to bottleneck our latents, we clamp them to a fixed numerical range and add diffusion-style noise.

\section{Related Work}

An existing body of work aims to generate 3D models by training auto-encoders on explicit 3D representations and then training generative models in the resulting latent space. \namecite{pcautoencoder} train an auto-encoder on point clouds, and experiment with both GANs \shortcite{gan} and GMMs \shortcite{gmmtraining} to model the resulting latent space. \namecite{pointflow} likewise train a point cloud auto-encoder, but their decoder is itself a conditional generative model (i.e. a normalizing flow \shortcite{flows}) over individual points in the point cloud; they also employ normalizing flows to model the latent space. \namecite{flowdiff} explore a similar technique, but use a diffusion model for the decoder instead of a normalizing flow. \namecite{lion} train a hierarchical auto-encoder, where the second stage encodes a point cloud of latent vectors instead of a single latent code; they employ diffusion models at both stages of the hierarchy. \namecite{textcraft} train a two-stage vector quantized auto-encoder \shortcite{vqvae,vqvae2} on voxel occupancy grids, and model the resulting latent sequences autoregressively. Unlike our work, these approaches all rely on explicit output representations which are often bound to a fixed resolution or lack the ability to fully express a 3D asset.

More similar to our own method, some prior works have explored 3D auto-encoders with implicit decoders. \namecite{shapecrafter} encode grids of SDF samples into latents which are used to condition an implicit SDF model. \namecite{clipforge} encode voxel grids into latents which are used to condition an implicit occupancy network. \namecite{towardsimplicit} train a voxel-based encoder and separate implicit occupancy and color decoders. \namecite{nerfvae} encode rendered views of a scene into latent vectors of a VAE, and this latent vector is used to condition a NeRF. Most similar to our encoder setup, \namecite{inr-encoder} use a transformer-based architecture to directly produce the parameters of an MLP conditioned on rendered views. We extend this prior body of work with \modelname{}, which produces more expressive implicit representations and is trained at a larger scale than most prior work.

While the above methods all train both encoders and decoders, other works aim to produce latent-conditional implicit 3D representations without a learned encoder. \namecite{deepsdf} train what they call an ``auto-decoder'', which uses a learned table of embedding vectors for each example in the dataset. In their case, they train an implicit SDF decoder that conditions on these per-sample latent vectors. \namecite{gaudi} uses a similar strategy to learn per-scene latent codes to condition a NeRF decoder. \namecite{functa} employ meta-learning to encode dataset examples as implicit functions. In their setup, they ``encode'' an example into (a subset of) the parameters of an implicit function by taking gradient steps on a reconstruction objective.
Concurrently to our work, \namecite{hyperdiffusion} utilize diffusion to directly generate the implicit MLP weights; however, akin to \shortcite{functa}, their method requires fitting NeRF parameters for each scene through gradient-based optimization.
\namecite{rodin} pursue a related approach, jointly training separate NeRFs for every sample in a dataset, but share a subset of the parameters to ensure that all resulting models use an aligned representation space. 
These approaches have the advantage that they do not require an explicit input representation. However, they can be expensive to scale with increasing dataset size, as each new sample requires multiple gradient steps. Moreover, this scalability issue is likely more pronounced for methods that do not incorporate meta-learning.

Several methods for 3D generation use gradient-based optimization to produce individual samples, often in the form of an implicit function. DreamFields \shortcite{dreamfields} optimizes the parameters of a NeRF to match a text prompt according to a CLIP-based \shortcite{clip} objective. DreamFusion \shortcite{dreamfusion} is a similar method with a different objective based on the output of a text-conditional image diffusion model. \namecite{magic3d} extend DreamFusion by optimizing a mesh representation in a second stage, leveraging the fact that meshes can be rendered more efficiently at higher resolution. \namecite{scorechaining} propose a different approach for leveraging text-to-image diffusion models, using them to optimize a differentiable 3D voxel grid rather than an MLP-based NeRF. While most of these approaches optimize implicit functions, \namecite{clipmesh} optimize the numerical parameters of a mesh itself, starting from a spherical mesh and gradually deforming it to match a text prompt. One common shortcoming of all of these approaches is that they require expensive optimization procedures, and a lot of work must be repeated for every sample that is generated. This is in contrast to direct generative models, which can potentially amortize this work by pre-training on a large dataset.

\section{Method}

In our method, we first train an encoder to produce implicit representations, and then train diffusion models on the latent representations produced by the encoder. Our method proceeds in two steps:

\begin{enumerate}
    \item We train an encoder to produce the parameters of an implicit function given a dense explicit representation of a known 3D asset (Section \ref{sec:3dencoder}). In particular, the encoder produces a latent representation of a 3D asset which is then linearly projected to obtain weights of a multi-layer perceptron (MLP).
    \item We train a diffusion prior on a dataset of latents obtained by applying the encoder to our dataset (Section \ref{sec:latentdiffusion}). This model is conditioned on either images or text descriptions.
\end{enumerate}

We train all of our models on a large dataset of 3D assets with corresponding renderings, point clouds, and text captions (\mbox{Section \ref{sec:dataset}}).

\subsection{Dataset}
\label{sec:dataset}

For most of our experiments, we employ the same dataset of underlying 3D assets as \namecite{pointe}, allowing for fairer comparisons with their method. However, we slightly extend the original post-processing as follows:

\begin{itemize}
  \item For computing point clouds, we render 60 views of each object instead of 20. We found that using only 20 views sometimes resulted in small cracks (due to blind spots) in the inferred point clouds.
  \item We produce point clouds of 16K points instead of 4K.
  \item When rendering views for training our encoder, we simplify the lighting and materials. In particular, all models are rendered with a fixed lighting configuration that only supports diffuse and ambient shading. This makes it easier to match the lighting setup with a differentiable renderer.
\end{itemize}

For our text-conditional model and the corresponding \pointe{} baseline, we employ an expanded dataset of underlying 3D assets and text captions. For this dataset, we collected roughly 1 million more 3D assets from high-quality data sources. Additionally, we gathered 120K captions from human labelers for high-quality subsets of our dataset. During training of our text-to-3D models, we randomly choose between human-provided labels and the original text captions when both are available.

\subsection{3D Encoder}
\label{sec:3dencoder}

\begin{figure}[t]
    \centering
    
    \includegraphics[width=1.0\textwidth]{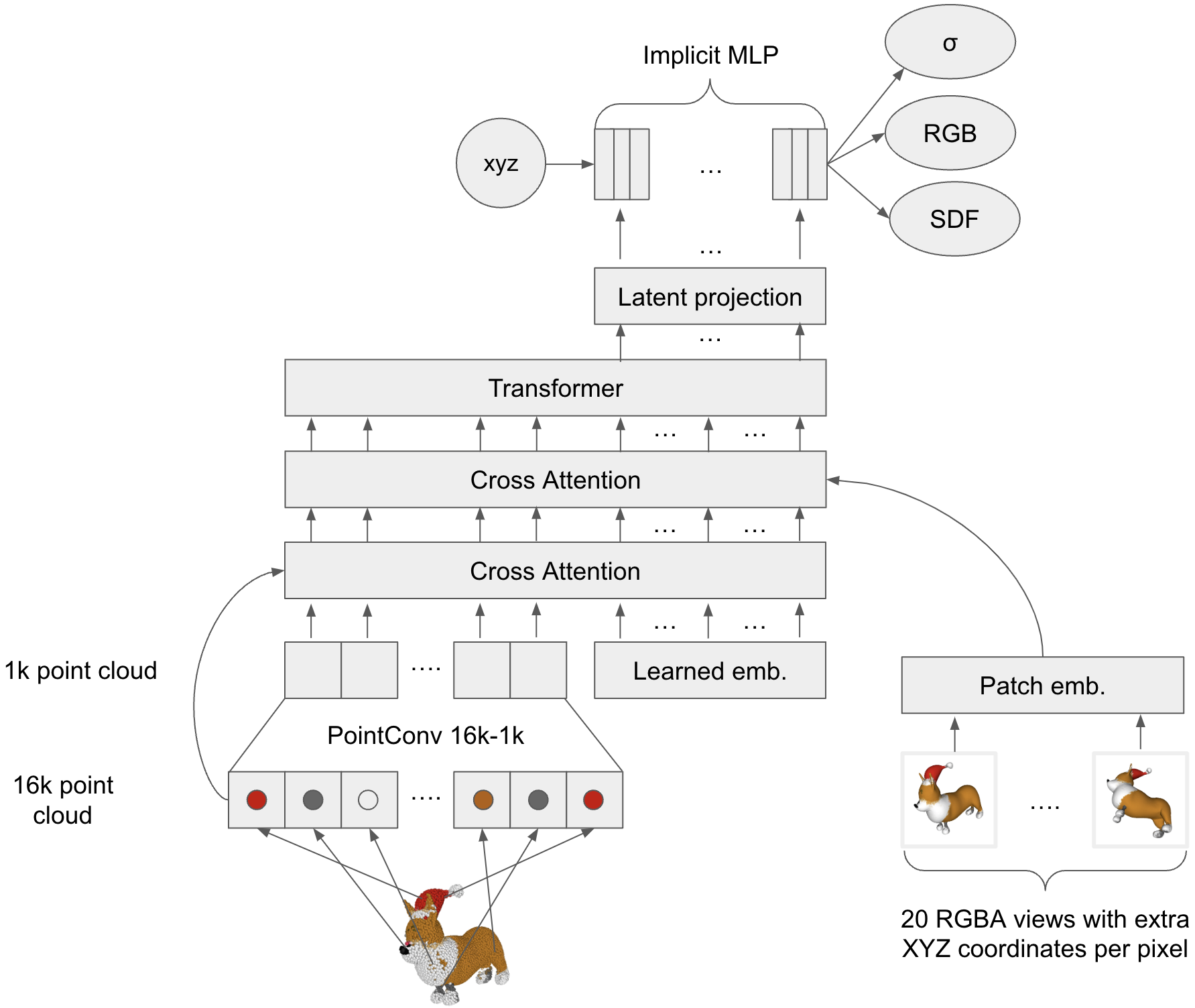}
    \caption{\label{fig:encoder} An overview of our encoder architecture. The encoder ingests both 16k resolution RGB point clouds and rendered RGBA images with augmented spatial coordinates for each foreground pixel. It outputs parameters of an MLP, which then acts as both a NeRF and a signed texture field (STF).}
\end{figure}

Our encoder architecture is visualized in Figure \ref{fig:encoder}. We feed the encoder both point clouds and rendered views of a 3D asset, and it outputs the parameters of a multi-layer perceptron (MLP) that represents the asset as an implicit function. Both the point cloud and input views are processed via cross-attention, which is followed by a transformer backbone that produces latent representations as a sequence of vectors. Each vector in this sequence is then passed through a latent bottleneck and projection layer whose output is treated as a single row of the resulting MLP weight matrices. During training, the MLP is queried and the outputs are used in either an image reconstruction loss or a distillation loss. For more details, see Appendix \ref{app:encoderarchitecture}.

\begin{figure*}[t]
    \centering
    \setlength{\tabcolsep}{2.0pt}
    \begin{tabular}{ccccccc}
        \scriptsize Ground-truth &
        \scriptsize \makecell{Pre-trained \\ NeRF} &
        \scriptsize \makecell{Pre-trained STF \\ (untrained)} &
        \scriptsize Distilled NeRF &
        \scriptsize Distilled STF &
        \scriptsize Finetuned NeRF &
        \scriptsize Finetuned STF \\
    
        \includegraphics[width=0.12\textwidth]{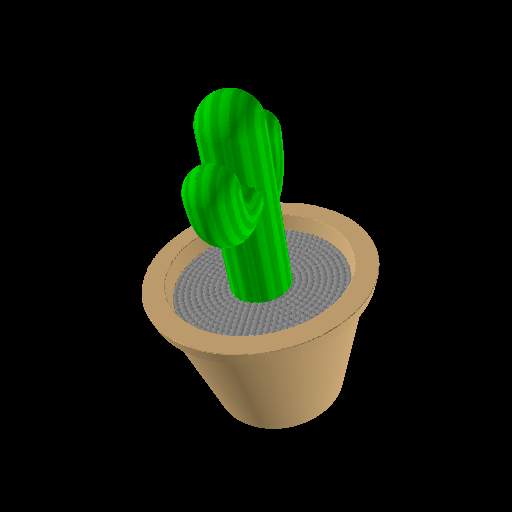} &
        \includegraphics[width=0.12\textwidth]{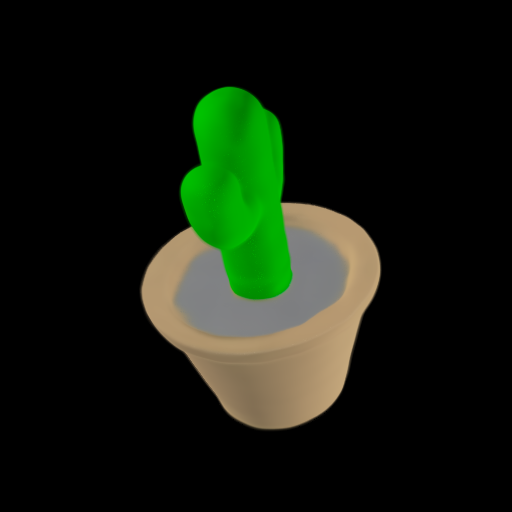} &
        \includegraphics[width=0.12\textwidth]{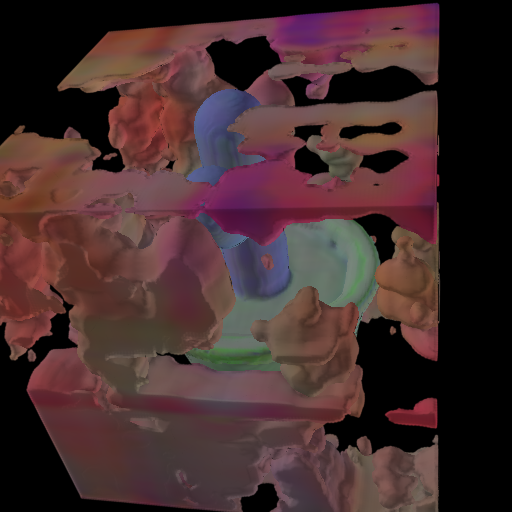} &
        \includegraphics[width=0.12\textwidth]{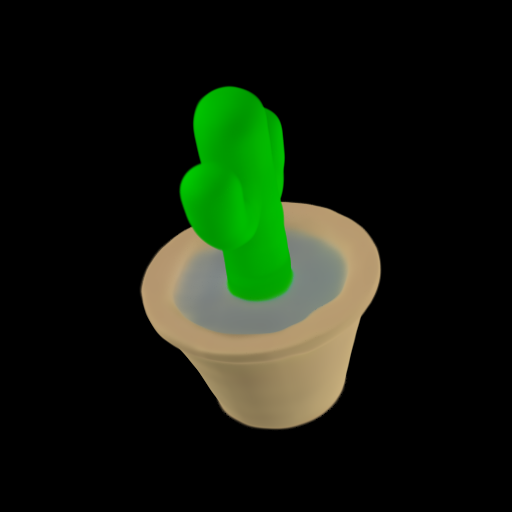} &
        \includegraphics[width=0.12\textwidth]{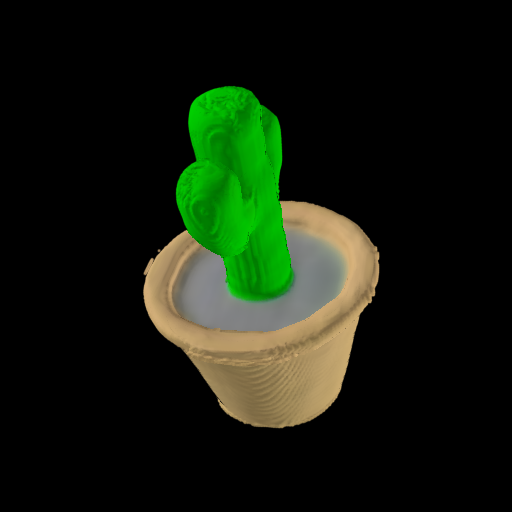} &
        \includegraphics[width=0.12\textwidth]{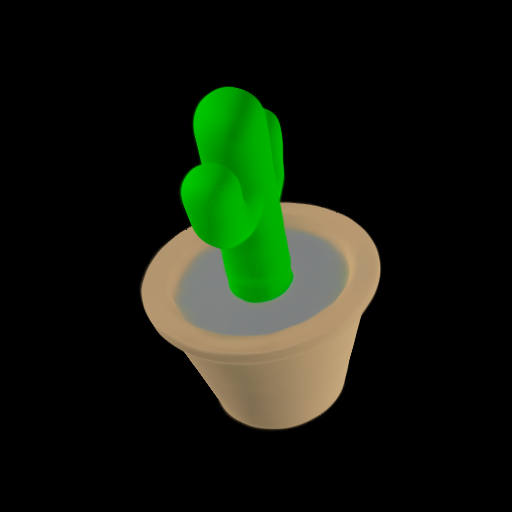} &
        \includegraphics[width=0.12\textwidth]{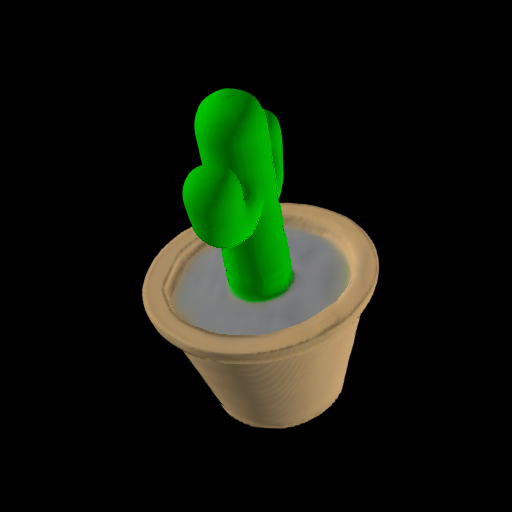} \\
        
        \includegraphics[width=0.12\textwidth]{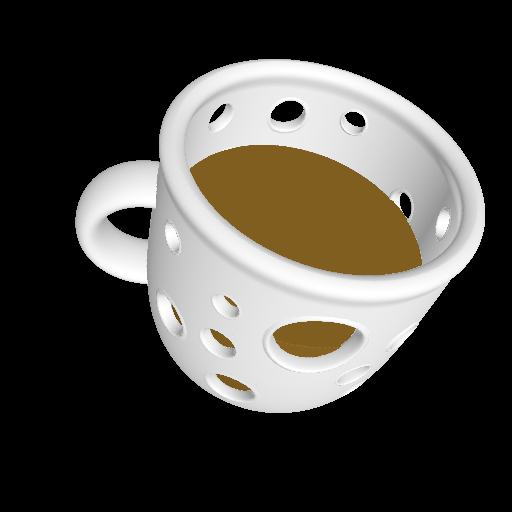} &
        \includegraphics[width=0.12\textwidth]{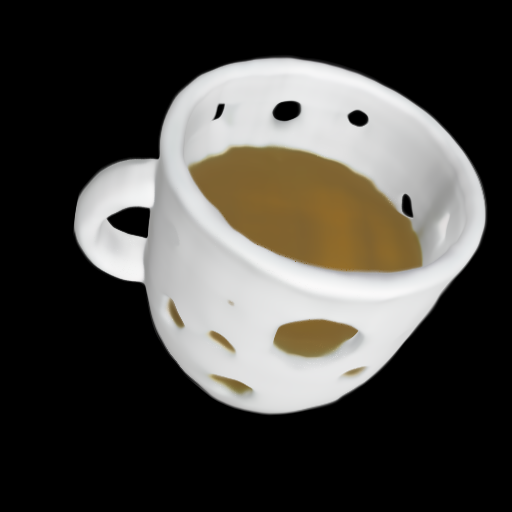} &
        \includegraphics[width=0.12\textwidth]{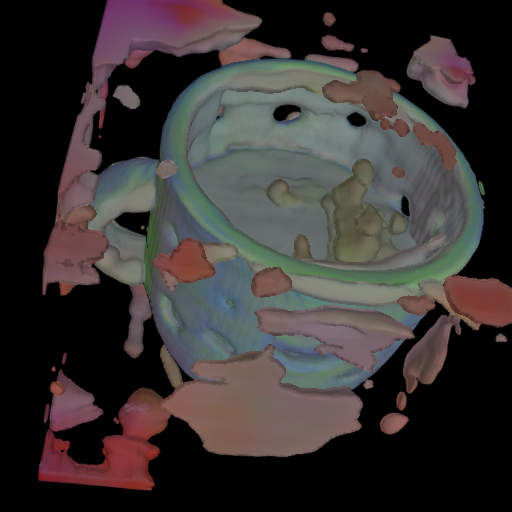} &
        \includegraphics[width=0.12\textwidth]{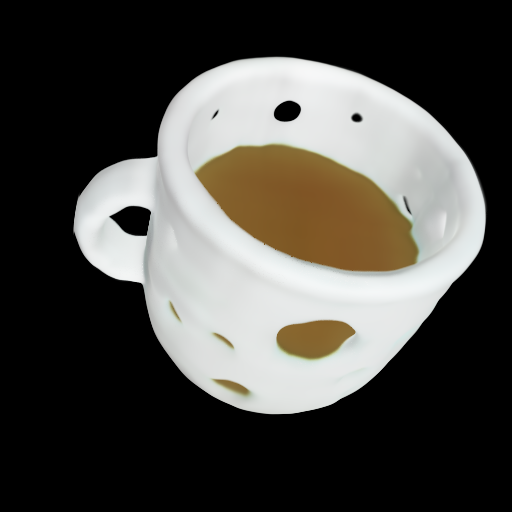} &
        \includegraphics[width=0.12\textwidth]{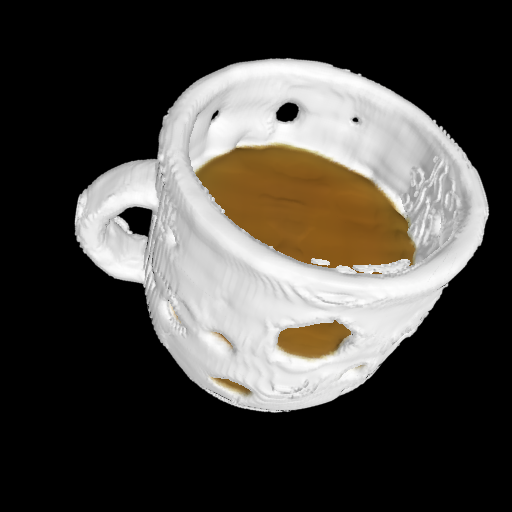} &
        \includegraphics[width=0.12\textwidth]{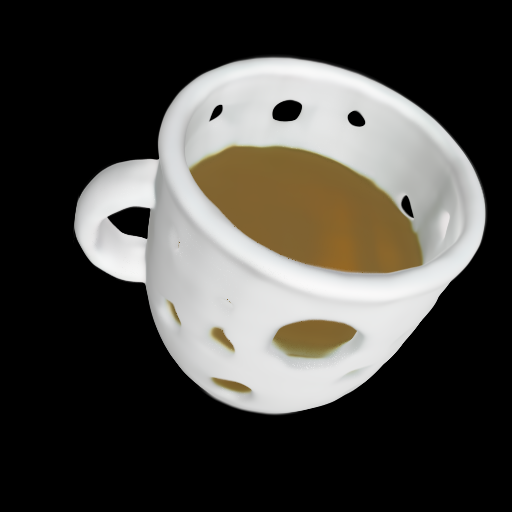} &
        \includegraphics[width=0.12\textwidth]{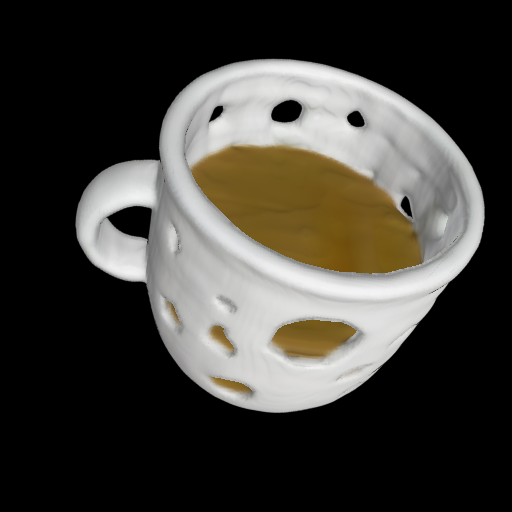}
        \rule{0pt}{0.15pt} \\
    \end{tabular}

    \caption{3D asset reconstructions from different rendering modes and checkpoints. Surprisingly, we find that randomly initialized STF heads still produce some elements of the original shape, likely because the previous layer activations are used for NeRF outputs. While distillation improves STF rendering results, it produces rough looking objects. Fine-tuning on both rendering methods yields the best reconstructions.}
    \label{fig:reconstructions}
    \vskip -0.1in
\end{figure*}

We pre-train our encoder using only a NeRF rendering objective (Section \ref{sec:nerfrendering}), as we found this to be more stable to optimize than mesh-based objectives. After NeRF pre-training, we add additional output heads for SDF and texture color predictions, and train these heads using a two-stage process (Section \ref{sec:stfrendering}). We show reconstructions of 3D assets for various checkpoints of our encoder with both rendering methods in Figure \ref{fig:reconstructions}.

\subsubsection{Decoding with NeRF Rendering}
\label{sec:nerfrendering}

We mostly follow the original NeRF formulation \shortcite{nerf}, except that we share the parameters between the coarse and fine models.\footnote{We use different linear output heads to produce coarse and fine predictions.} We randomly sample 4096 rays for each training example, and minimize an $L_1$ loss\footnote{In preliminary scans, we found that $L_1$ loss outperformed $L_2$ loss on PSNR after an initial warmup period where $L_1$ was worse.} between the true color $C(\mathbf{r})$ and the predicted color from the NeRF:
$$L_\text{RGB} = E_{\mathbf{r} \in R} \bigl[||\hat{C}_c(\mathbf{r}) - C(\mathbf{r})||_1 + ||\hat{C}_f(\mathbf{r}) - C(\mathbf{r})||_1\bigr]$$

We also add an additional loss on the transmittance of each ray. In particular, the integrated density of a ray gives transmittance estimates $\hat{T}_c(r)$ and $\hat{T}_f(r)$ for coarse and fine rendering, respectively. We use the alpha channel from the ground-truth renderings to obtain transmittance targets $T(r)$, giving a second loss:
$$L_{T} = E_{\mathbf{r} \in R} \bigl[ ||\hat{T}_c(\mathbf{r}) - T(\mathbf{r})||_1 + ||\hat{T}_f(\mathbf{r}) - T(\mathbf{r})||_1 \bigr]$$

We then optimize the joint objective:
$$L_{\text{NeRF}} = L_\text{RGB} + L_{T}$$

\subsubsection{Decoding with STF Rendering}
\label{sec:stfrendering}

After NeRF-only pre-training, we add additional STF output heads to our MLPs which predict SDF values and texture colors. To construct a triangle mesh, we query the SDF at vertices along a regular $128^3$ grid and apply a differentiable implementation of Marching Cubes 33 \shortcite{machingcubes33}. We then query the texture color head at each vertex of the resulting mesh. We differentiably render the resulting textured mesh using PyTorch3D \shortcite{pytorch3d}. We always render with the same (diffuse) lighting configuration which is identical to the lighting configuration used to preprocess our dataset.

In preliminary experiments, we found that randomly-initialized STF output heads were unstable and difficult to train with a rendering-based objective. To alleviate this issue, we first distill approximations of the SDF and texture color into these output heads before directly training with differentiable rendering. In particular, we randomly sample input coordinates and obtain SDF distillation targets using the \pointe{} SDF regression model, and RGB targets using the color of the nearest neighbor in the asset's RGB point cloud. During distillation training, we use a sum of distillation losses and the pre-training NeRF loss:

$$L_\text{distill} = L_{\text{NeRF}} + E_{\mathbf{x} \sim U[-1, 1]^3}\bigl[ ||\text{SDF}_{\theta}(\mathbf{x}) - \text{SDF}_{\text{regression}}(\mathbf{x})||_1 
 + ||\text{RGB}_{\theta}(\mathbf{x}) - \text{RGB}_{\text{NN}}(\mathbf{x})||_1 \bigr]$$

Once the STF output heads have been initialized to reasonable values via distillation, we fine-tune the encoder for both NeRF and STF rendering end-to-end. We found it unstable to use $L_1$ loss for STF rendering, so we instead use $L_2$ loss only for this rendering method. In particular, we optimize the following loss for STF rendering:
$$L_\text{STF} = \frac{1}{N \cdot s^2} \sum_{i=1}^{N} ||\textrm{Render}(\text{Mesh}_i) - \textrm{Image}_{i}||_2^2$$

where $N$ is the number of images, $s$ is the image resolution, $\text{Mesh}_i$ is the constructed mesh for sample $i$, $\text{Image}_i$ is a target RGBA rendering for image $i$, and $\text{Render}(x)$ renders a mesh using a differentiable renderer. We do not include a separate transmittance loss, since this is already captured by the alpha channel of the image.

For this final fine-tuning step, we optimize the summed objective:
$$L_\text{FT} = L_{\text{NeRF}} + L_\text{STF}$$

\subsection{Latent Diffusion}
\label{sec:latentdiffusion}

For our generative models, we adopt the transformer-based diffusion architecture of \pointe{}, but replace point clouds with sequences of latent vectors. Our latents are sequences of shape $1024 \times 1024$, and we feed this into the transformer as a sequence of $1024$ tokens where each token corresponds to a different row of the MLP weight matrices. As a result, our models are roughly compute equivalent to the base \pointe{} models (i.e. have the same context length and width) while generating samples in a much higher-dimensional space due to the increase of input and output channels.

We follow the same conditioning strategies as \pointe{}. For image-conditional generation, we prepend a 256-token CLIP embedding sequence to the Transformer context. For text-conditional generation, we prepend a single token containing the CLIP text embedding. To support classifier-free guidance, we randomly set the conditioning information to zero during training with probability 0.1.

Unlike \pointe{}, we do not parameterize our diffusion model outputs as $\epsilon$ predictions. Instead, we directly predict $x_0$, which is algebraically equivalent to predicting $\epsilon$, but produced more coherent samples in early experiments. The same observation was made by \namecite{unclip}, who opted to use $x_0$ prediction when generating CLIP latent vectors with diffusion models.

\section{Results}

\subsection{Encoder Evaluation}

We track two render-based metrics throughout the encoder training process. First, we evaluate the peak signal-to-noise ratio (PSNR) between reconstructions and ground-truth rendered images. Additionally, to measure our encoder's ability to capture semantically relevant details of 3D assets, we encode meshes produced by the largest \pointe{} model and re-evaluate the CLIP R-Precision of the reconstructed NeRF and STF renders. Table \ref{tab:encodereval} tracks these two metrics over the different stages of training. We find that distillation hurts NeRF reconstruction quality, but fine-tuning recovers (and slightly boosts) NeRF quality while drastically increasing the quality of STF renders. 

\begin{table}[t]
    \caption{\label{tab:encodereval} Evaluating the encoder after each stage of training. We evaluate PSNR between reconstructions and ground-truth renders, as well as CLIP R-Precision on reconstructions of samples from \pointe{} 1B (where the peak performance is roughly $46.8\%$).}
    \centering
    \begin{center}
    \begin{small}
    \begin{tabular}{ccccc}
    \toprule
    Stage & NeRF PSNR (dB) & STF PSNR (dB) & \makecell{NeRF \pointe{} CLIP \\ R-Precision} & \makecell{STF \pointe{} CLIP \\ R-Precision} \\
    \midrule
    Pre-training (300K) & 33.2 & -    & 44.3\% & - \\
    Pre-training (600K) & 34.5 & -    & 45.2\% & - \\
    Distillation        & 32.9 & 23.9 & 42.6\% & 41.1\% \\
    Fine-tuning         & 35.4 & 31.3 & 45.3\% & 44.0\% \\
    \bottomrule
    \end{tabular}
    \end{small}
    \end{center}
    \vskip -0.2in
\end{table}

\subsection{Comparison to \texorpdfstring{\pointe{}}{Point-E}}

\begin{figure}[t]
    \centering
    \includegraphics[width=0.9\textwidth]{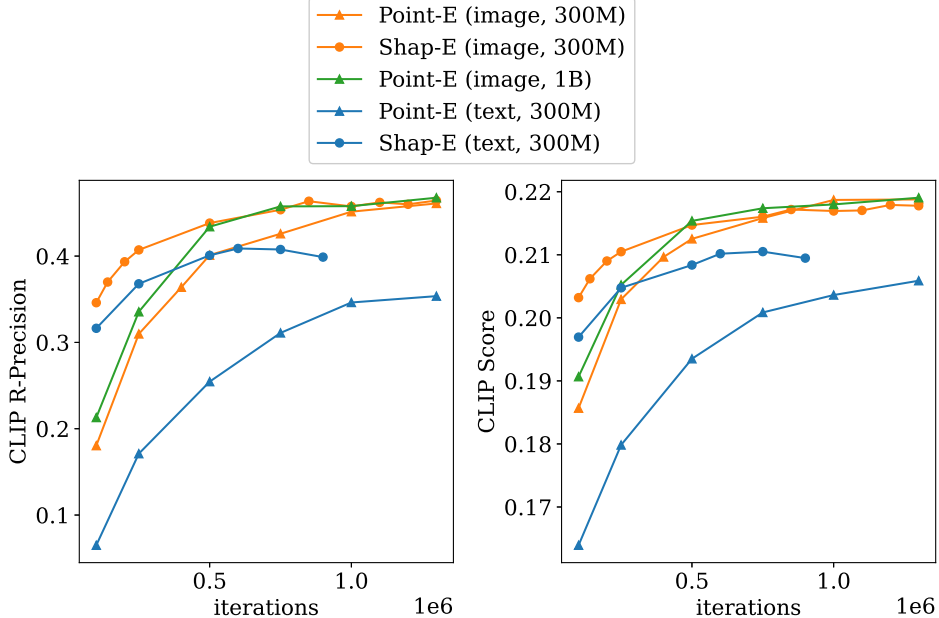}
    \caption{\label{fig:pointetrainingplots}Evaluations throughout training for both \modelname{} and \pointe{}. For each checkpoint for both models, we take the maximum value when sweeping over guidance scales $\{2.0, 3.0, 4.0, 5.0, 8.0, 10.0, 15.0\}$.}
\end{figure}

Our latent diffusion model shares the same architecture, training dataset, and conditioning modes as \pointe{}.\footnote{However, note that \modelname{} depends on a separate encoder model, while \pointe{} depends on separate upsampler and SDF models. Only the base diffusion model architecture is the same.} As a result, comparing to \pointe{} helps us isolate the effects of generating implicit neural representations rather than an explicit representation. We compare these methods throughout training on sample-based evaluation metrics in Figure \ref{fig:pointetrainingplots}. As done by \namecite{dreamfields} and various follow-up literature, we compute CLIP R-Precision \shortcite{rprecision} on a set of COCO validation prompts. We also evaluate CLIP score on these same prompts, as this metric is often used for measuring image generation quality \shortcite{glide}. We only train comparable 300 million parameter models, but we also plot evaluations for the largest (1 billion parameter) \pointe{} model for completeness.

\begin{figure}[t]
    \centering
    \begin{tabular}{ccc}
        \scriptsize \makecell{Prompt} &
        \scriptsize \pointe{} Samples &
        \scriptsize \modelname{} Samples (ours) \\
        
        \raisebox{0.033\textwidth}{``a diamond ring''} &
        \includegraphics[width=0.35\textwidth]{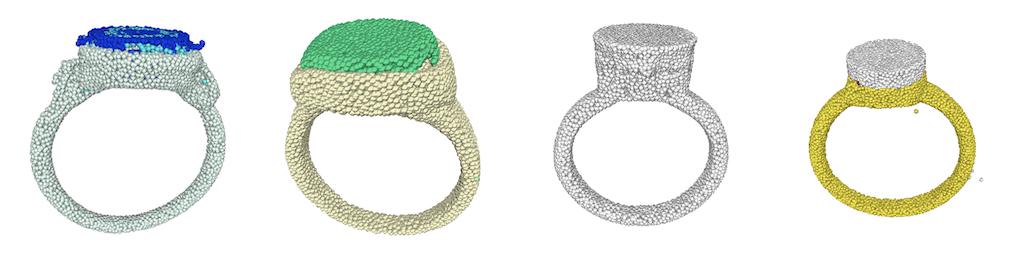} &
        \includegraphics[width=0.35\textwidth]{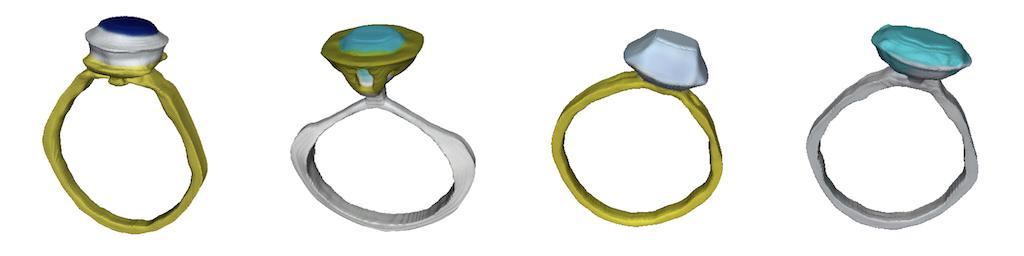} \\

        \raisebox{0.033\textwidth}{``a traffic cone''} &
        \includegraphics[width=0.35\textwidth]{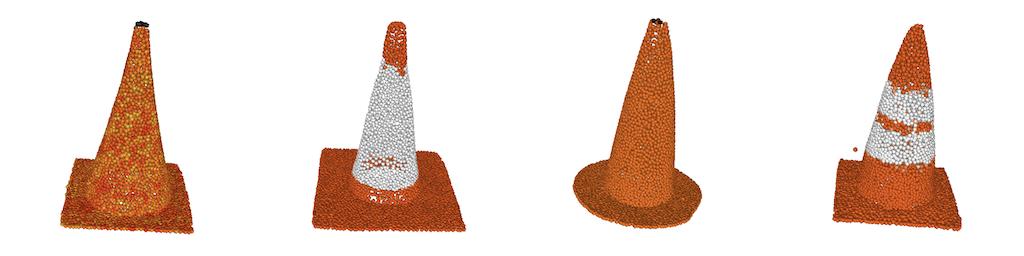} &
        \includegraphics[width=0.35\textwidth]{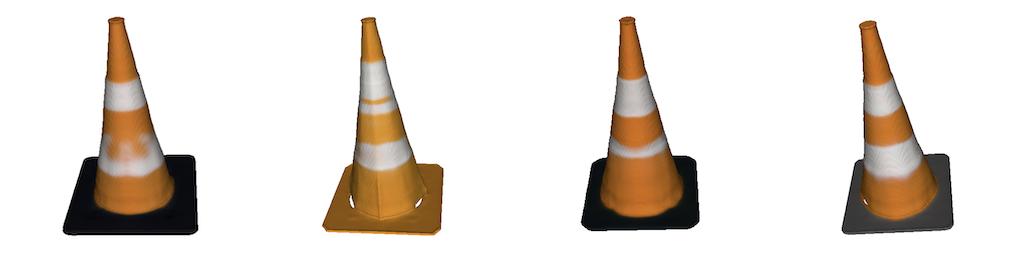} \\

        \raisebox{0.04\textwidth}{\makecell{``a donut with \\ pink icing''}} &
        \includegraphics[width=0.35\textwidth]{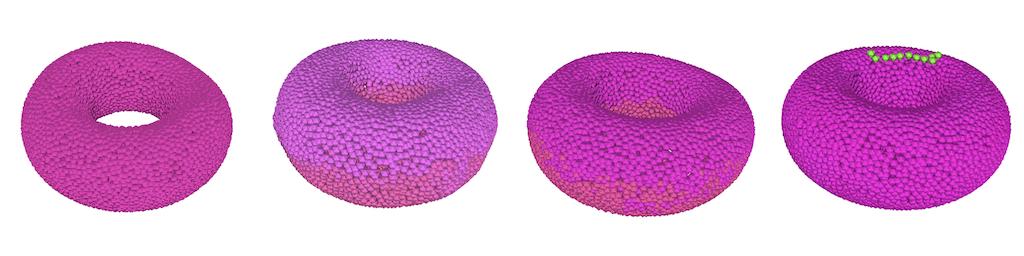} &
        \includegraphics[width=0.35\textwidth]{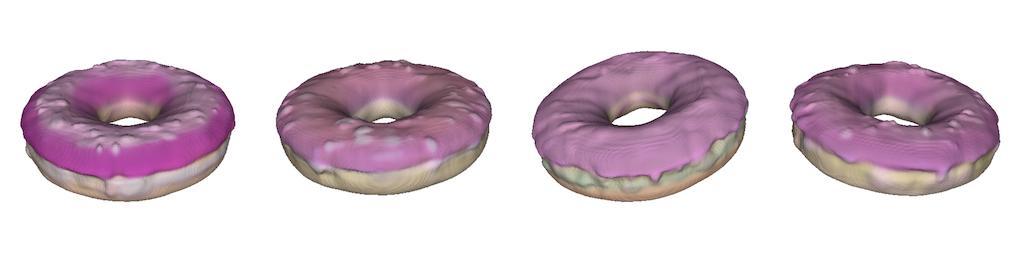} \\

        \raisebox{0.033\textwidth}{``a corgi''} &
        \includegraphics[width=0.35\textwidth]{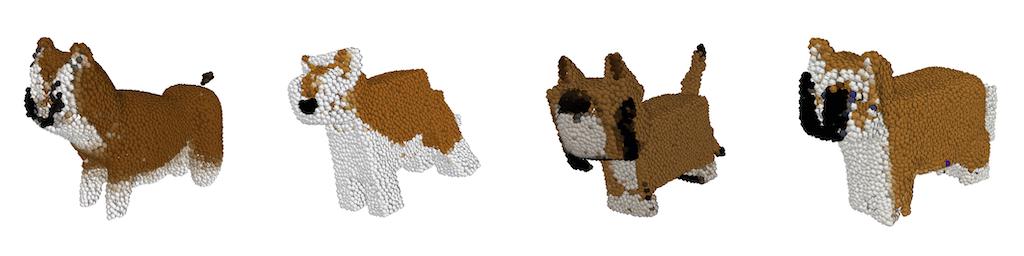} &
        \includegraphics[width=0.35\textwidth]{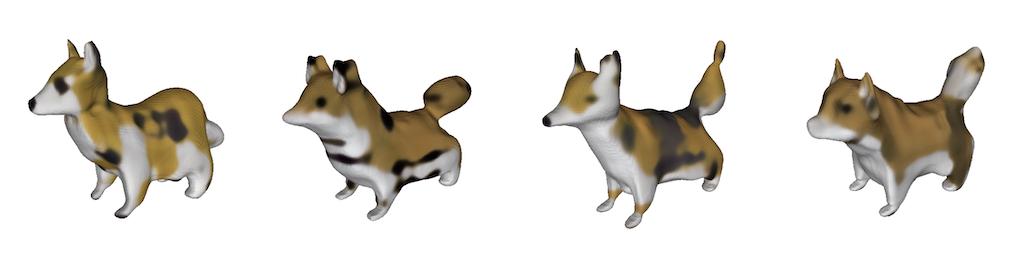} \\

        \raisebox{0.033\textwidth}{``a designer dress''} &
        \includegraphics[width=0.35\textwidth]{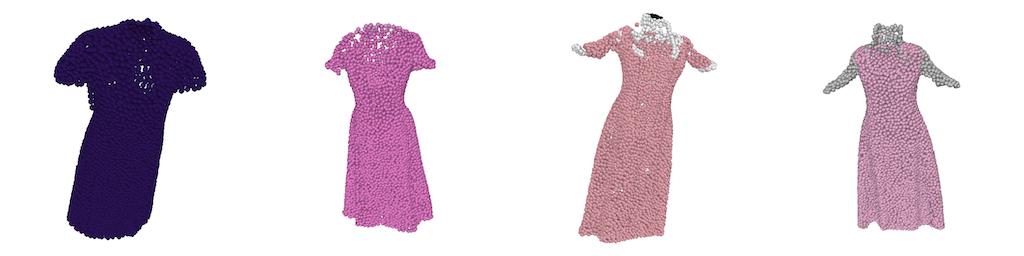} &
        \includegraphics[width=0.35\textwidth]{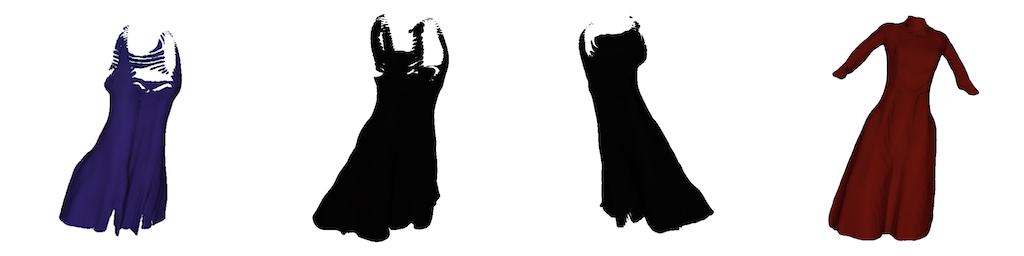} \\

        \raisebox{0.033\textwidth}{``a pair of shorts''} &
        \includegraphics[width=0.35\textwidth]{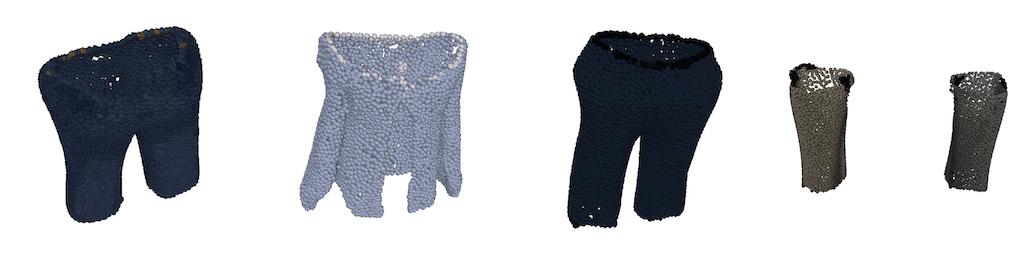} &
        \includegraphics[width=0.35\textwidth]{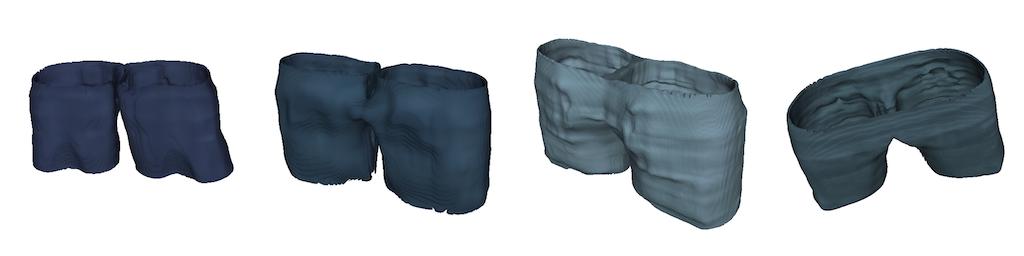} \\

        \raisebox{0.033\textwidth}{``a hypercube''} &
        \includegraphics[width=0.35\textwidth]{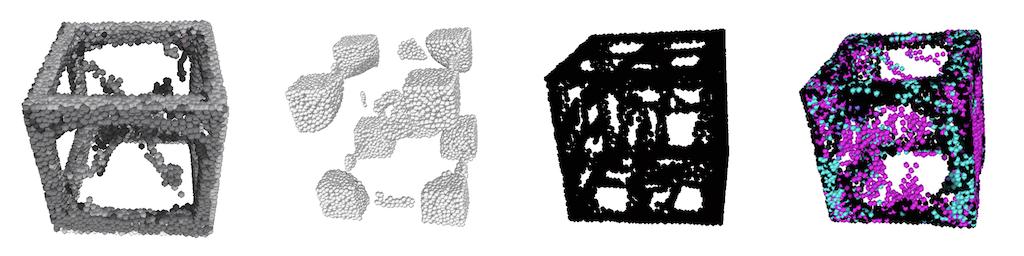} &
        \includegraphics[width=0.35\textwidth]{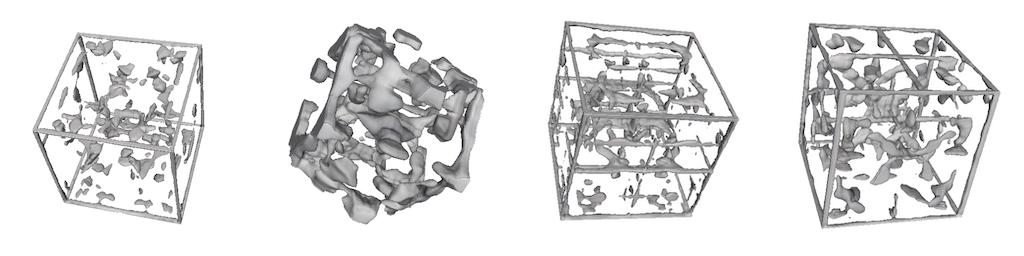} \\
    \end{tabular}

    \caption{\label{fig:textqualitative} Examples of text prompts for which text-conditional \pointe{} and \modelname{} consistently exhibit qualitatively different behavior. For each prompt, we show four random samples from both models, which were trained on the same dataset with the same base model size.}
\end{figure}

In the text-conditional setting, we observe that \modelname{} improves on both metrics over the comparable \pointe{} model. To rule out the possibility that this gap is due to perceptually small differences, we also show qualitative samples in Figure \ref{fig:textqualitative}, finding that these models often produce qualitatively different samples for the same text prompts. We also observe that our text-conditional \modelname{} begins to get worse on evaluations before the end of training. In Appendix \ref{app:overfitting}, we argue that this is likely due to overfitting to the text captions, and we use an early-stopped checkpoint for all figures and tables.

\begin{figure}[t]
    \centering
    \begin{subfigure}[b]{0.8\textwidth}
        \centering
        \begin{tabular}{ccc}
            \scriptsize \makecell{Conditioning \\ image} &
            \scriptsize \pointe{} Sample &
            \scriptsize \modelname{} Sample \\
            \includegraphics[width=0.3\textwidth]{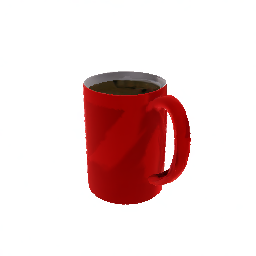} &
            \includegraphics[width=0.3\textwidth]{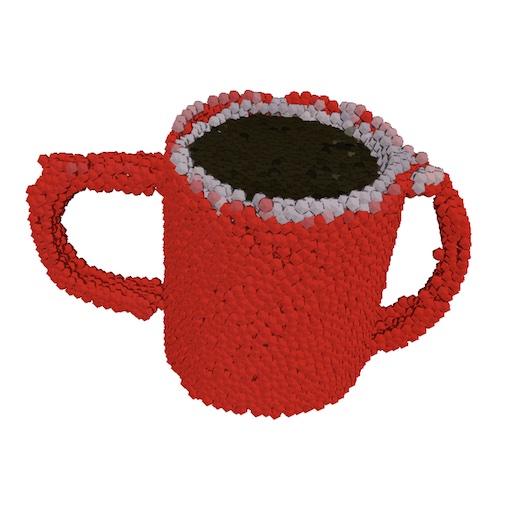} &
            \includegraphics[width=0.3\textwidth]{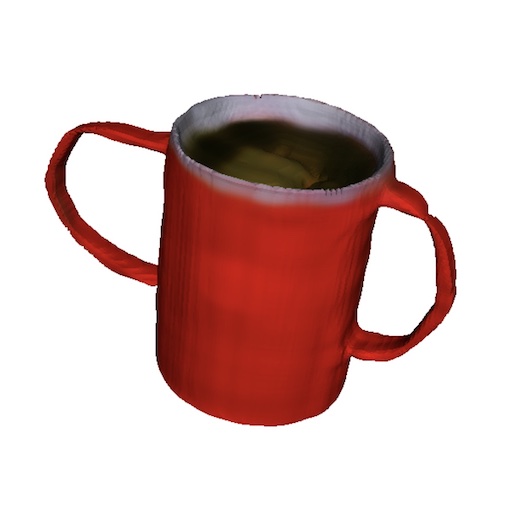} \\
    
            \includegraphics[width=0.3\textwidth]{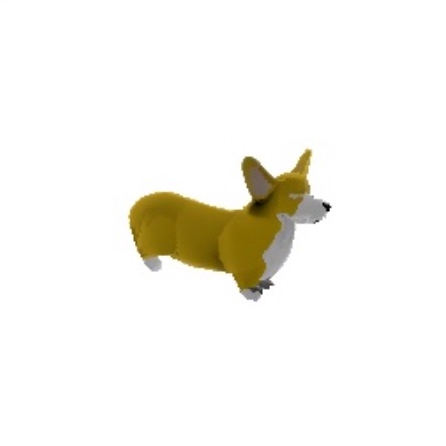} &
            \includegraphics[width=0.3\textwidth]{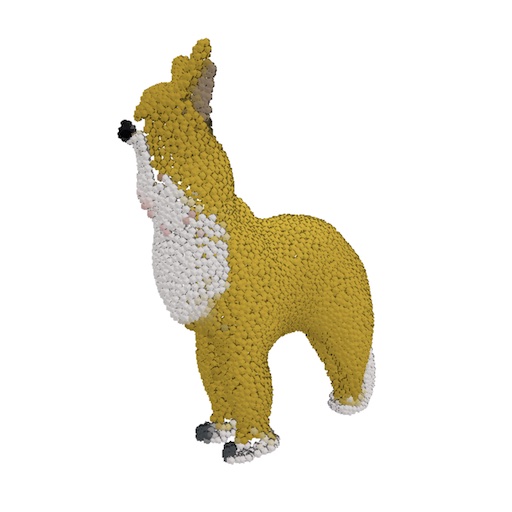} &
            \includegraphics[width=0.3\textwidth]{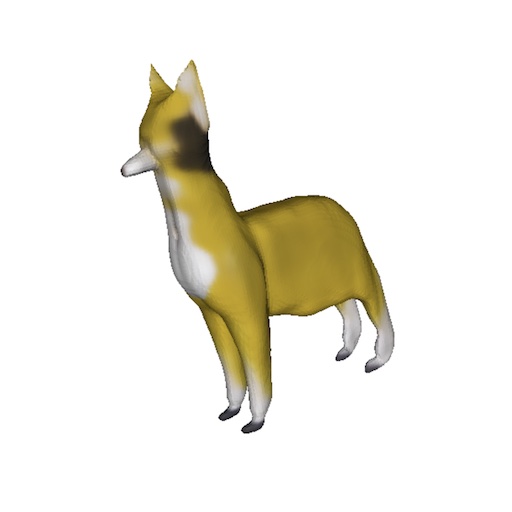} \\
        \end{tabular}

        \caption{\label{fig:sharedfailures} Shared failure cases between image-conditional \modelname{} and \pointe{}. In the first example, both models counter-intuitively infer an occluded handle on the mug. In the second, both models incorrectly interpret the proportions of the depicted animal.}
    \end{subfigure}
    \begin{subfigure}[b]{0.8\textwidth}
        \centering
        \vspace{10pt}
        \begin{tabular}{ccc}
            \scriptsize \makecell{Conditioning \\ image} &
            \scriptsize \pointe{} Sample &
            \scriptsize \modelname{} Sample \\

            \includegraphics[width=0.3\textwidth]{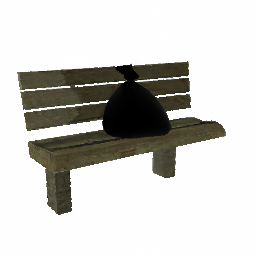} &
            \includegraphics[width=0.3\textwidth]{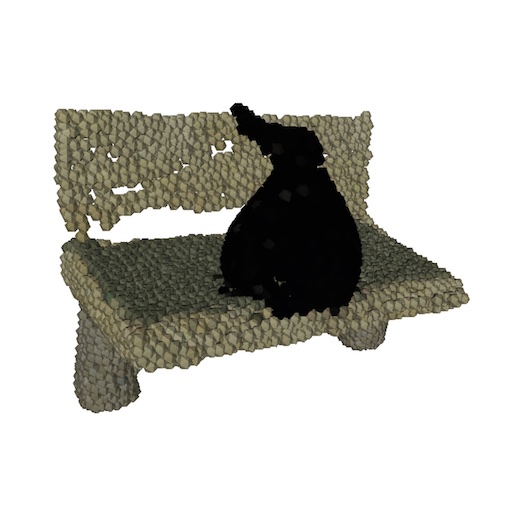} &
            \includegraphics[width=0.3\textwidth]{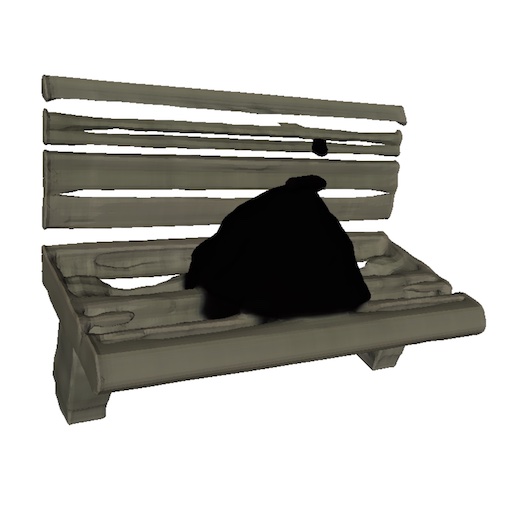} \\

            \includegraphics[width=0.3\textwidth]{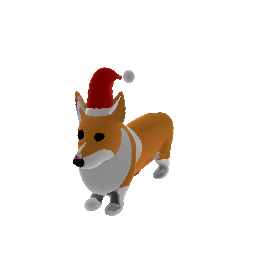} &
            \includegraphics[width=0.3\textwidth]{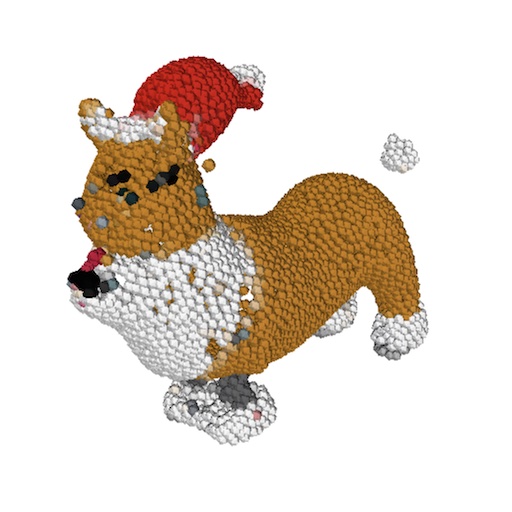} &
            \includegraphics[width=0.3\textwidth]{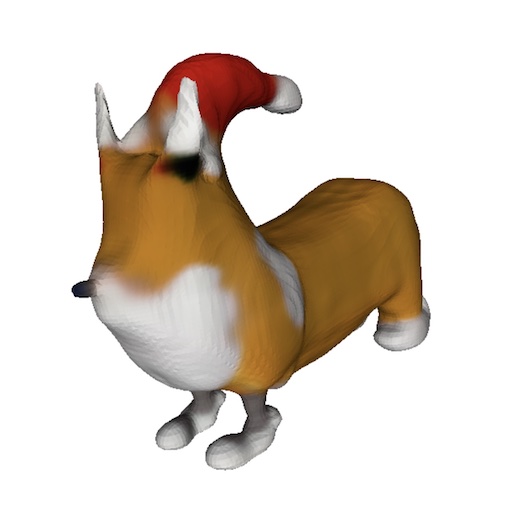} \\
        \end{tabular}
        \caption{\label{fig:sharedsuccesses} Conditioning images for which both \modelname{} and \pointe{} succeed.}
    \end{subfigure}

    \caption{Randomly selected image-conditional samples from both \pointe{} and \modelname{} for the same conditioning images.}
    \vskip -0.1in
\end{figure}

Unlike the text-conditional case, our image-conditional \modelname{} and \pointe{} models reach roughly the same final evaluation performance, with a slight advantage for \modelname{} in CLIP R-Precision and a slight disadvantage in CLIP score. To investigate this phenomenon more deeply, we inspected samples from both models. We initially expected to see qualitatively different behavior from the two models, since they produce samples in different representation spaces. However, we discovered that both models tend to share similar failure cases, as shown in Figure \ref{fig:sharedfailures}. This suggests that the training data, model architecture, and conditioning images affect the resulting samples more than the chosen representation space.

However, we do still observe some qualitative differences between the two image-conditional models. For example, in the first row of Figure \ref{fig:sharedsuccesses}, we find that \pointe{} sometimes ignores the small slits in the bench, whereas \modelname{} attempts to model them. We hypothesize that this particular difference could occur because point clouds are a poor representation for thin features or gaps.
Also, we observe in Table \ref{tab:encodereval} that the 3D encoder slightly reduces CLIP R-Precision when applied to \pointe{} samples. Since \modelname{} achieves comparable CLIP R-Precision as \pointe{}, we hypothesize that \modelname{} must generate qualitatively different samples for some prompts which are not bottlenecked by the encoder. This further suggests that explicit and implicit modeling can still learn distinct features from the same data and model architecture.

\subsection{Comparison to Other Methods}

We compare \modelname{} to a broader class of 3D generative techniques on the CLIP R-Precision metric in Table \ref{tab:cliprprec}. As done by \namecite{pointe}, we include sampling latency in this table to highlight that the superior sample quality of optimization-based methods comes at a significant inference cost. We also note that \modelname{} enjoys faster inference than \pointe{} because \modelname{} does not require an additional upsampling diffusion model.

\begin{table}[t]
    \caption{\label{tab:cliprprec} Comparison of 3D generation techniques on the CLIP R-Precision metric on COCO evaluation prompts. Compute estimates and other methods' values are taken from \namecite{pointe}. $^*$The best text-conditional results are obtained using our expanded dataset of 3D assets.}
    \vskip 0.15in
    \centering
    \begin{center}
    \begin{small}
    \begin{tabular}{rccl}
    \toprule
    Method & ViT-B/32 & ViT-L/14 & Latency \\
    \midrule
    DreamFields & 78.6\% & 82.9\% & $\sim 200$ V100-hr \\
    CLIP-Mesh & 67.8\% & 74.5\% & $\sim 17$ V100-min \\
    DreamFusion & 75.1\% & 79.7\% & $\sim 12$ V100-hr \\
    \midrule
    \makecell{\pointe{} (300M, text-only)} & 33.6\%$^*$ & 35.5\%$^*$ & 25 V100-sec \\
    \makecell{\modelname{} (300M, text-only)} & 37.8\%$^*$ & 40.9\%$^*$ & 13 V100-sec \\
    \midrule
    \pointe{} (300M) & 40.3\% & 45.6\% & 1.2 V100-min \\
    \pointe{} (1B) & 41.1\% & 46.8\% & 1.5 V100-min \\
    \modelname{} (300M) & 41.1\% & 46.4\% & 1.0 V100-min \\
    \midrule
    \midrule
    \makecell[r]{Conditioning \\ images} & 69.6\% & 86.6\% & - \\
    \bottomrule
    \end{tabular}
    \end{small}
    \end{center}
    \vskip -0.2in
\end{table}

\section{Limitations and Future Work}

\begin{figure}[t]
    \centering
    \begin{tabular}{ccc}
        \scriptsize \makecell{Prompt} &
        \scriptsize Text-conditional samples \\

        \raisebox{0.05\textwidth}{\makecell{``a stool with a green \\ seat and red legs''}} &
        \includegraphics[width=0.45\textwidth]{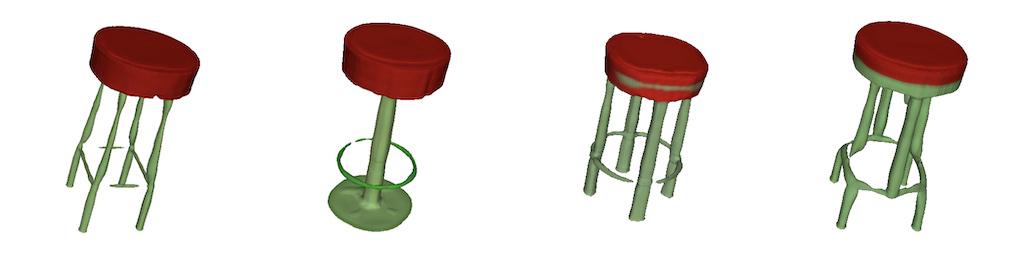} \\

        \raisebox{0.05\textwidth}{\makecell{``a red cube on top of \\ a blue cube''}} &
        \includegraphics[width=0.45\textwidth]{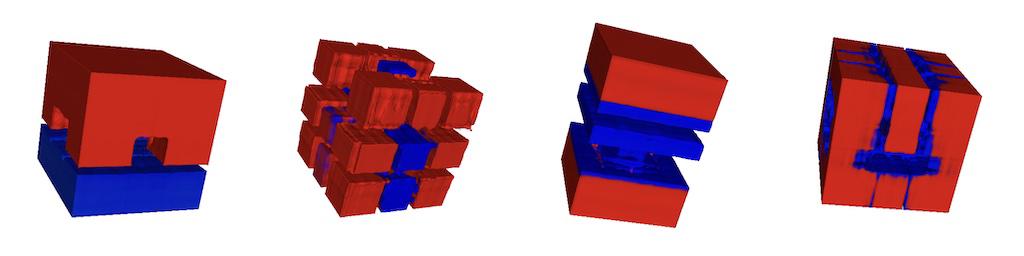} \\

        \raisebox{0.05\textwidth}{\makecell{``two cupcakes''}} &
        \includegraphics[width=0.45\textwidth]{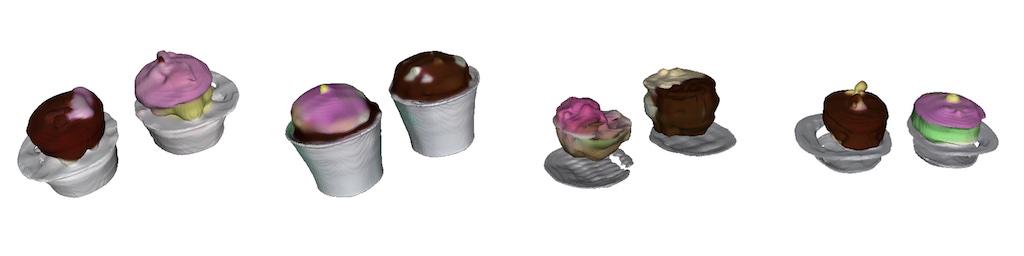} \\

        \raisebox{0.05\textwidth}{\makecell{``three cupcakes''}} &
        \includegraphics[width=0.45\textwidth]{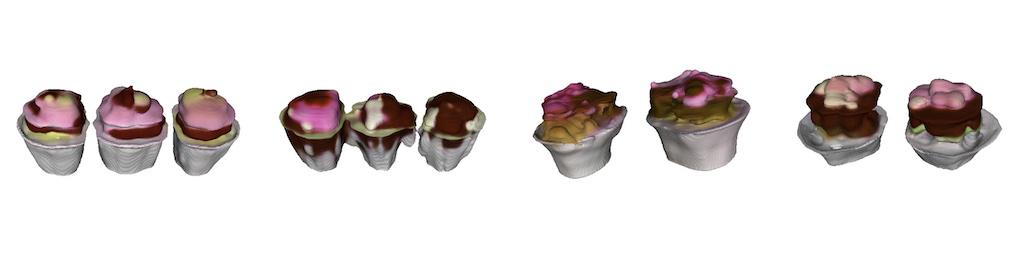} \\

        \raisebox{0.05\textwidth}{\makecell{``four cupcakes''}} &
        \includegraphics[width=0.45\textwidth]{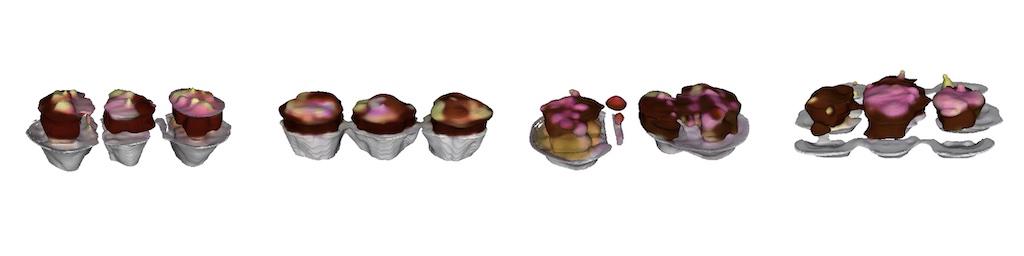} \\
    \end{tabular}

    \caption{\label{fig:textfailures} Examples of text-conditional \modelname{} samples prompts which require counting and attribute binding.}
\end{figure}

While our text-conditional model can understand many single object prompts with simple attributes, it has a limited ability to compose concepts. In Figure \ref{fig:textfailures}, we find that this model struggles to bind multiple attributes to different objects, and fails to reliably produce the correct number of objects when asked for more than two. These failures are likely the result of limited paired training data, and could potentially be alleviated by gathering or generating larger annotated 3D datasets.

Additionally, while \modelname{} can often produce recognizable 3D assets, the resulting samples often look rough or lack fine details. Notably, Figure \ref{fig:reconstructions} shows that the encoder itself sometimes loses detailed textures (e.g. the stripes on the cactus), indicating that improved encoders could potentially recover some of the lost generation quality.

For the best results, \modelname{} could potentially be combined with optimization-based 3D generative techniques. For example, a NeRF or mesh produced by \modelname{} could be used to initialize an optimization-based approach such as DreamFusion, potentially leading to faster convergence. Alternatively, image-based objectives could be used to guide the \modelname{} sampling process, as we briefly explore in Appendix \ref{app:guideddreamfusion}.

\section{Conclusion}

We present \modelname{}, a latent diffusion model over a space of 3D implicit functions that can be rendered as both NeRFs and textured meshes. We find that \modelname{} matches or outperforms a similar explicit generative model given the same dataset, model architecture, and training compute. We also find that our pure text-conditional models can generate diverse, interesting objects without relying on images as an intermediate representation. These results highlight the potential of generating implicit representations, especially in domains like 3D where they can offer more flexibility than explicit representations.

\section{Acknowledgments}

Our thanks go to Prafulla Dhariwal, Joyce Lee, Jack Rae, and Mark Chen for helpful discussions, and to all contributors of ChatGPT, which provided valuable writing feedback.

\setcitestyle{numbers}
\bibliography{main}
\bibliographystyle{plainnat}

\clearpage

\appendix

\section{Hyperparameters}
\label{app:hyperparameters}

\subsection{Encoder Architecture}
\label{app:encoderarchitecture}

To capture details of the input 3D asset, we feed our encoder two separate representations of a 3D model:

\begin{itemize}
    \item \textbf{Point clouds:} For each 3D asset, we pre-compute an RGB point cloud with 16,384 points.
    \item \textbf{Multiview point clouds:} In addition to a point cloud, we render 20 views of each 3D asset from random camera angles at $256 \times 256$ resolution. We augment each foreground pixel with an $(x,y,z)$ surface coordinate, giving an image of shape $256 \times 256 \times 7$. We apply an $8 \times 8$ patch embedding these augmented renderings, resulting in a sequence of 20,480 vectors representing a \textit{multiview point cloud}.
\end{itemize}

Our encoder begins by using a point convolution layer to downsample the input point cloud into a set of 1K embeddings. This set of embeddings is concatenated with a learned input embedding $h_l$ to obtain a query sequence $h$. We then update $h$ with a single cross-attention layer that references the input point cloud. Next, we update $h$ again by cross-attending to the patch embedded multiview point cloud $m$. Next, we apply a transformer to $h$ and take the 1K suffix tokens as latent vectors. 
We then apply a $\textrm{tanh}(x)$ activation to these latents to clamp them to the range $[-1, 1]$. At this stage, we have obtained the latent vector that we target with our diffusion models, but we do not yet have the parameters of an MLP.

After computing the sequence of latents, we apply Gaussian diffusion noise $q(h_t)$ to the latents with probability 0.1. For this diffusion noise, we use the schedule $\bar{\alpha}_t = 1 - t^5$ which typically produces very little noise. After the noise and bottleneck layers, we project each latent vector to 256 dimensions and stack the resulting latents into four MLP weight matrices of size $256 \times 256$. Our full encoder architecture is described in Algorithm \ref{alg:encoder}.

\begin{algorithm}[t]
    \caption{\label{alg:encoder} High-level pseudocode of our encoder architecture.}
    \textbf{Input} point cloud $p$, multiview point cloud $m$, learned input embedding sequence $h_l$.
    
    \textbf{Outputs:} latent variable $h$ and MLP parameters $\theta$.
    
    \begin{algorithmic}[1]
    \State $h \gets \textrm{ Cat}([\textrm{PointConv}(p), h_l])$
    \State $h \gets \textrm{CrossAttend}(h, \textrm{Proj}(p))$
    \State $h \gets \textrm{CrossAttend}(h, \textrm{PatchEmb}(m))$
    \State $h \gets \textrm{Transformer}(h)$
    \State $h \gets h[-\textrm{len}(h_l):]$
    \State $h \gets \textrm{tanh}(h)$
    \State $h' \gets \textrm{DiffusionNoise}(h)$
    \State $\theta \gets \textrm{Proj}(h')$
    \State \Return $h, \theta$
    \end{algorithmic}
\end{algorithm}

\subsection{Encoder Training}

We pre-train our encoders for 600K iterations using Adam \shortcite{adam} with a learning rate of $10^{-4}$ and a batch size of 64. We perform STF distillation for 50K iterations with a learning rate of $10^{-5}$ and keep the batch size at 64. We query 32K random points on each 3D asset for STF distillation. We fine-tune on STF renders for 65K iterations with the same hyperparameters as for distillation. For each stage of training, we re-initialize the optimizer state. For pre-training, we use 16-bit precision with loss scaling \shortcite{lossscaling}, but we found full 32-bit precision necessary to stabilize fine-tuning.

\subsection{Implicit Representations}

We represent our INRs as 6-layer MLPs where the first four layers are determined by the output of an encoder; the final two layers are shared across dataset examples. We do not use any biases in these models. Input coordinates are concatenated with sinusoidal embeddings following the work of \namecite{nerf} and \namecite{3dim}. In particular, each coordinate dimension $x$ is expanded as

$$[x, \cos(2^0 x), \sin(2^0 x), ..., \cos(2^{14} x), \sin(2^{14} x)]$$

Our MLPs use SiLU activations \shortcite{silu} between intermediate layers. The NeRF density and RGB heads are followed by sigmoid and ReLU activations, respectively. The SDF and texture color heads are followed by tanh and sigmoid activations, respectively.

Although we use direction independent lighting in all of our experiments, we found our encoders unstable to train unless we augmented their input coordinates with ray direction embeddings. Unlike typical NeRF models, our models' density head can be influenced by the ray direction, potentially leading to view-inconsistent objects. To ensure view-consistency at test time, we always set the ray direction embeddings to zero. Despite the out-of-distribution inputs, this approach is effective, likely because the model learns to disregard the ray direction with sufficient training. It remains an open question why the ray direction is beneficial during initial pre-training, yet appears irrelevant in later stages of training.

\subsection{Diffusion Models}

\begin{figure*}[t]
    \centering
    \begin{tabular}{c}
        \includegraphics[width=0.8\textwidth]{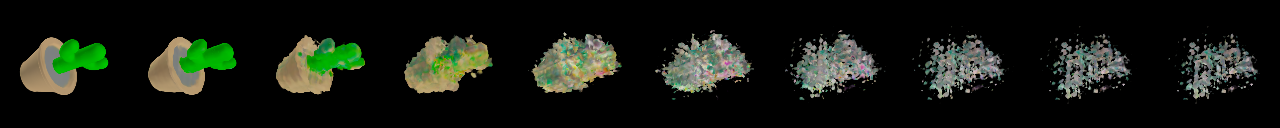} \\
        \includegraphics[width=0.8\textwidth]{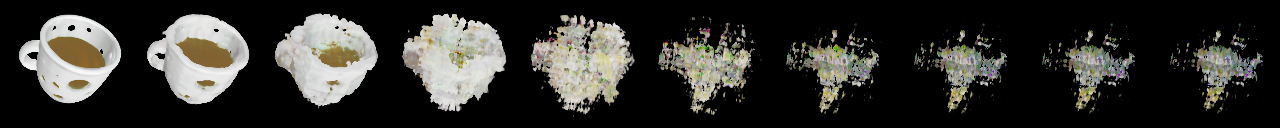}
    \end{tabular}

    \caption{NeRF reconstructions of noised latents using our diffusion schedule $\bar{\alpha}_t = e^{-12 t}$. The timestep $t$ is linearly swept from 0 to 1 from left to right.}
    \label{fig:noisingprocess}
\end{figure*}

When training our diffusion models, we employ the same hyperparameters as used for the 300M parameter \pointe{} models. The only difference is that we use larger input and output projections to accommodate 1024 feature channels (instead of 6).

For diffusion models, the choice of noise schedule $\bar{\alpha}_t$ can often have a big impact on sample quality \shortcite{improved,edm,perceptual}. Intuitively, the relative scale between a sample and the noise injected at a particular timestep determines how much information is destroyed at that timestep, and we would like a noise schedule that gradually destroys semantic information in the signal. In early experiments, we tried the cosine \shortcite{improved} and linear \shortcite{ddpm} schedules, as well as a schedule which we found visually to destroy information gradually: $\bar{\alpha}_t = e^{-12 t}$ (see Figure \ref{fig:noisingprocess}). In these experiments, we found that the latter schedule performed better on evaluation metrics, and decided to use it for all future experiments.

\begin{figure*}[t]
    \centering
    \includegraphics[width=0.5\textwidth]{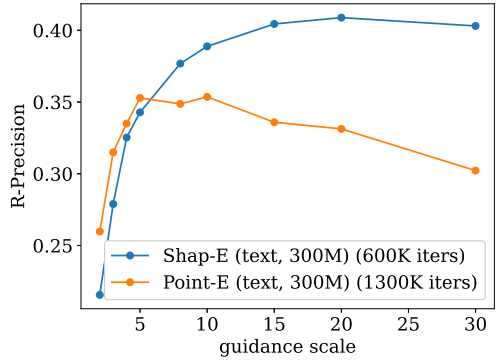}
    \caption{Evaluation sweep over guidance scale for text-conditional models. We find that \modelname{} benefits from increasing guidance scale up to 20.0, whereas \pointe{} begins to saturate at lower guidance scales and then becomes worse.}
    \label{fig:guidancescales}
\end{figure*}

We use similar Heun sampling hyperparameters as \pointe{}, but found that setting $s_\text{churn} = 0$ was a better choice for \modelname{}, whereas $s_\text{churn} = 3$ was better for \pointe{}. Additionally, we found that, while our image-conditional models tended to prefer the same guidance scale as \pointe{}, our text-conditional models could tolerate much higher guidance scales while still improving on evaluations (Figure \ref{fig:guidancescales}). We find that our best text-conditional \pointe{} samples are obtained using a scale of 5.0, while the best \modelname{} results use a scale of 20.0.

\subsection{Evaluation}

When evaluating CLIP-based metrics, we render our models' samples using NeRF at $128 \times 128$ resolution. We sample camera positions randomly around the z-axis, with a constant 30 degree elevation for all camera poses. We find that this works well in practice, since objects in our training dataset are usually oriented with the z-axis as the logical vertical direction.

\section{Overfitting in Text-Conditional Models}
\label{app:overfitting}

\begin{figure}[t]
    \centering
    \begin{subfigure}[b]{0.4\textwidth}
        \centering
        \includegraphics[width=\textwidth]{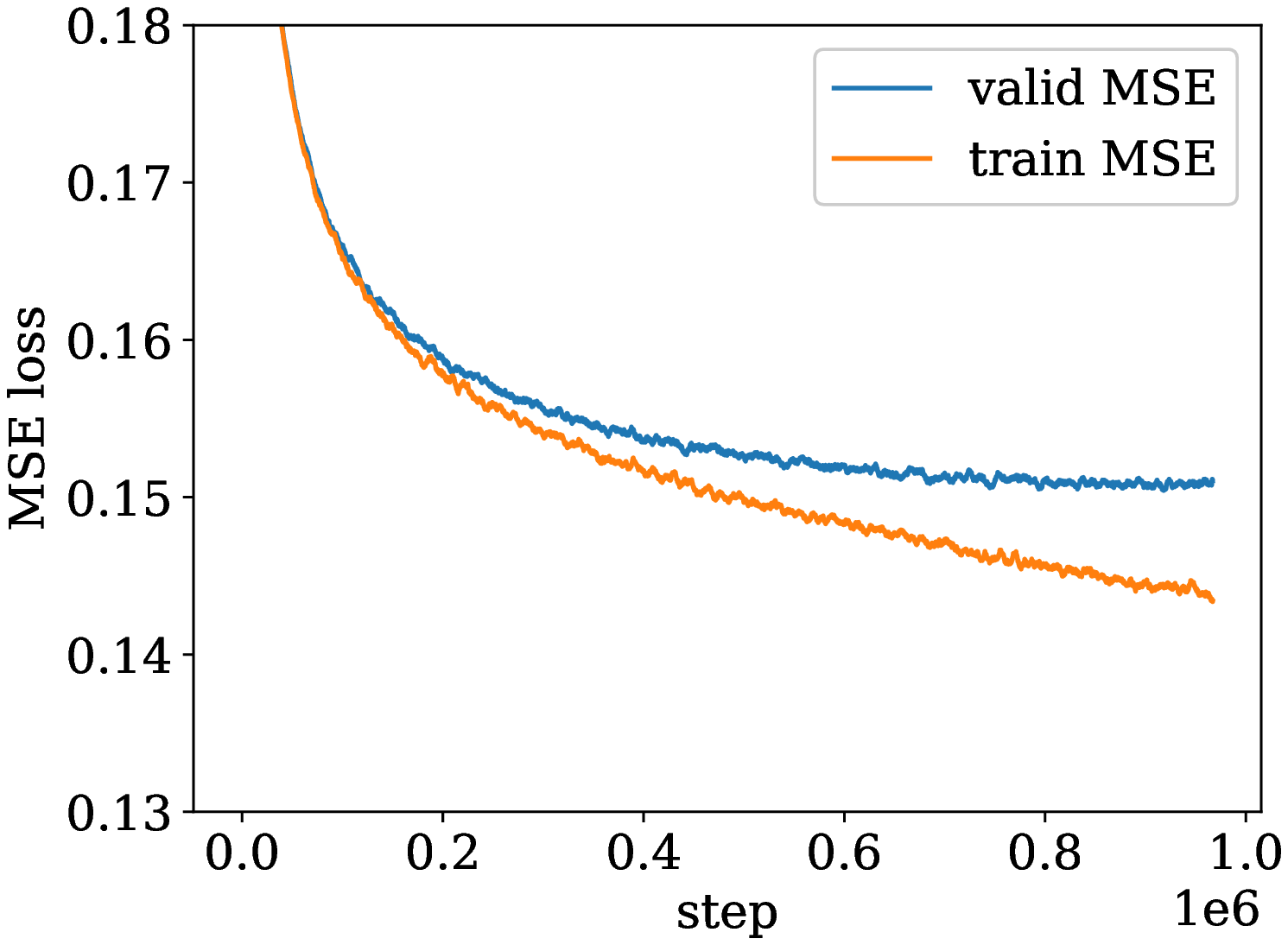}
        \caption{\label{fig:overfittingplot} Train and validation loss averaged across all diffusion steps.}
    \end{subfigure}
    \hspace{0.075\textwidth}
    \begin{subfigure}[b]{0.4\textwidth}
        \centering
        \includegraphics[width=\textwidth]{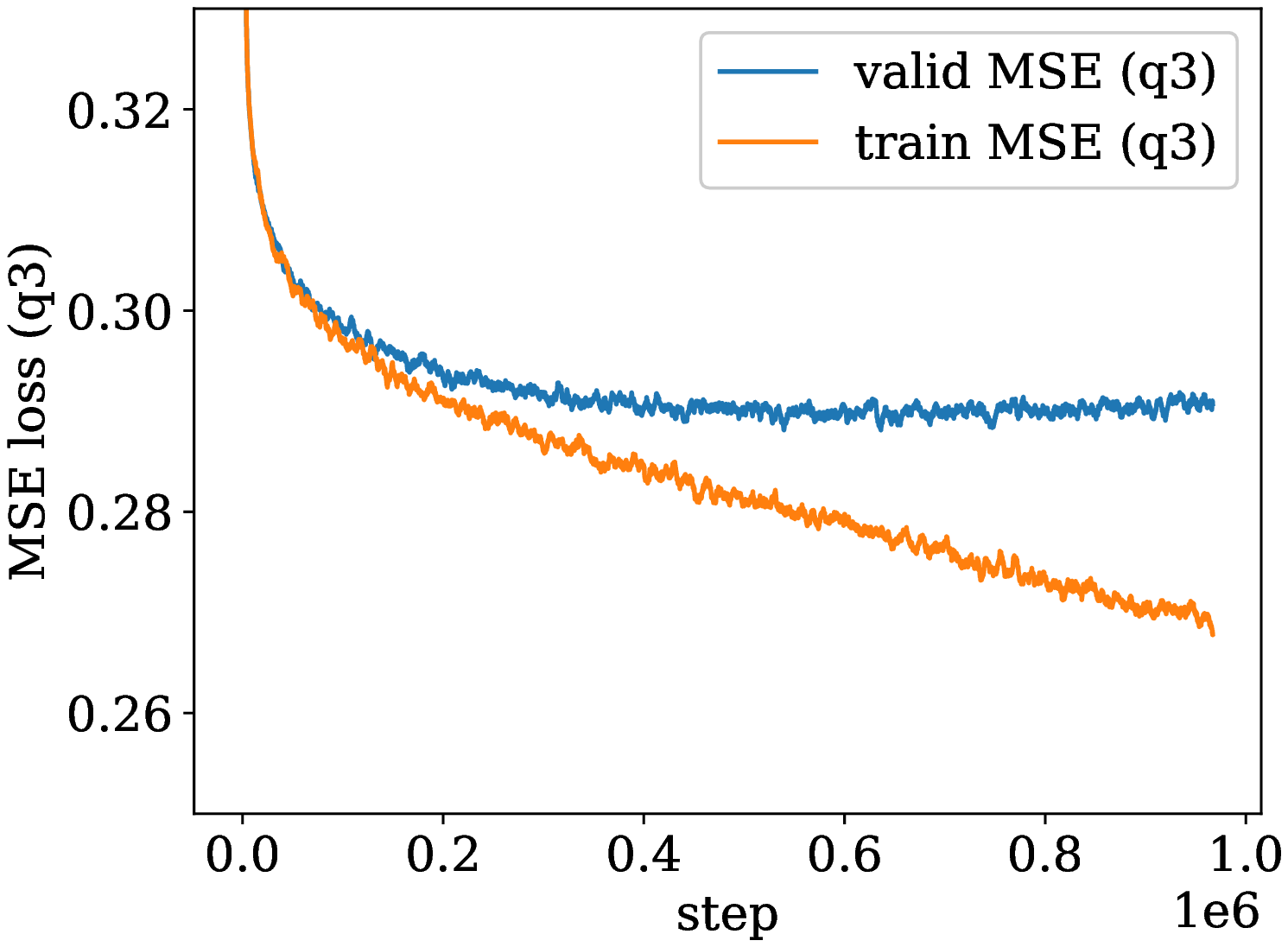}
        \caption{\label{fig:overfittingplotq3} Train/validation loss averaged over the noisiest quarter of the diffusion steps.}
    \end{subfigure}
    \caption{\label{fig:overfittingplots} Training and validation losses for our text-conditional model. We find that this model overfits, and that the overfitting is stronger for the noisiest diffusion timesteps.}
\end{figure}

We observe that our text-conditional model begins to get worse on evaluations after roughly 600K iterations. We hypothesize that this is due to overfitting to the text captions in the dataset, since we did not observe this phenomenon in the image-conditional case. In Figure \ref{fig:overfittingplot}, we observe that the training loss decreases faster than the validation loss, but that validation loss itself never starts increasing. Why, then, does the model get worse on evaluations?

To more deeply explore this overfitting, we leverage the fact that the diffusion loss is actually a sum of many different loss terms at different noise levels. In Figure \ref{fig:overfittingplotq3}, we plot the training and validation losses over only the noisiest quarter of the diffusion steps, finding that in this case overfitting is more pronounced and the validation loss indeed starts increasing at about 600K iterations. Intuitively, conditioning information is more likely to affect noisier timesteps since less information can be inferred from the noised sample $x_t$. This supports the hypothesis that the overfitting is tied to the model's understanding of the conditioning signal, although it may still be overfitting to other aspects of the data.

\section{Bias and Misuse}

\begin{figure}[t]
    \centering
    \begin{tabular}{cc}
        \scriptsize \makecell{Prompt} &
        \scriptsize Samples \\
        
        \raisebox{0.04\textwidth}{``a doctor''} &
        \includegraphics[width=0.4\textwidth]{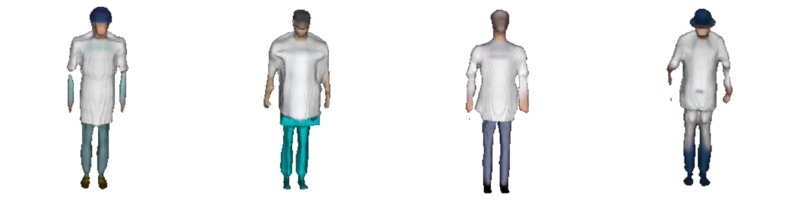} \\

        \raisebox{0.04\textwidth}{``a nurse''} &
        \includegraphics[width=0.4\textwidth]{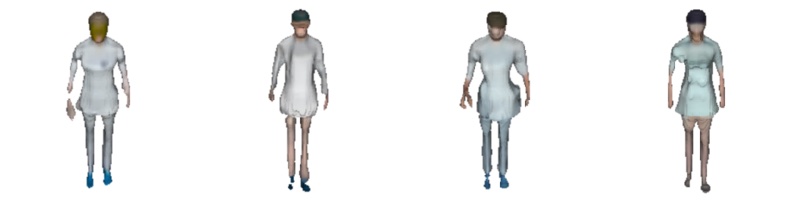} \\

        \raisebox{0.04\textwidth}{``an engineer''} &
        \includegraphics[width=0.4\textwidth]{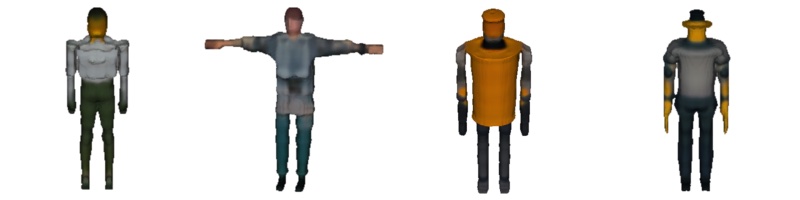} \\
    \end{tabular}

    \caption{\label{fig:biasexamples} Examples where our text-conditional model likely exhibits biases from its dataset.}
\end{figure}

Biases present in our dataset are likely to impact the behavior of the models we develop. In Figure \ref{fig:biasexamples}, we examine bias within our text-conditional model by providing it with ambiguous captions in which certain details, such as body shape or color, are left unspecified. We observe that the samples generated by the model exhibit common gender-role stereotypes in response to these ambiguous prompts.

\begin{figure}[t]
    \centering
    \begin{tabular}{cc}
        \scriptsize \makecell{Prompt} &
        \scriptsize Samples \\
        
        \raisebox{0.04\textwidth}{``a 1/8" titanium drill bit''} &
        \includegraphics[width=0.4\textwidth]{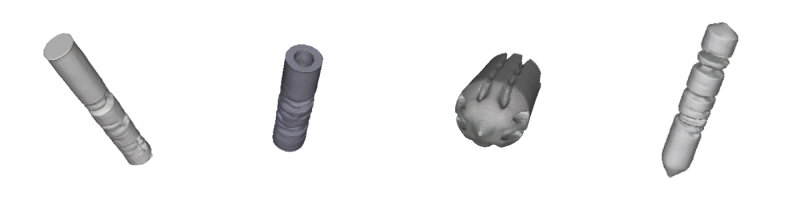} \\

    \end{tabular}

    \caption{\label{fig:misuseexamples} Examples of generated 3D objects which could have adverse consequences if used in the real world without validation.}
\end{figure}

Our models are not typically adept at producing photo-realistic samples or accurately following long and complex prompts, and this limitation comes with both benefits and drawbacks. On the positive side, it alleviates concerns regarding the potential use of our models to create convincing ``DeepFakes'' \shortcite{deepfakes}. On the negative side, it raises potential risks when our models are used in conjunction with fabrication methods such as 3D printing to create tools and parts (e.g. Figure \ref{fig:misuseexamples}). In such scenarios, 3D objects generated by the model could be introduced into the real world without undergoing adequate validation or safety testing, and this could potentially be harmful when the produced samples do not adequately meet the desired prompt.

\section{Guidance in Image Space}
\label{app:guideddreamfusion}

\begin{figure}[t]
    \centering
    \begin{tabular}{cccc}
        \makecell{$s$} &
        \makecell{``a corgi''} &
        \makecell{``a corgi wearing \\ a santa hat''} &
        \makecell{``a red cube on top \\ of a blue cube''} \\
        
        \raisebox{0.05\textwidth}{$s=0$} &
        \includegraphics[width=0.15\textwidth]{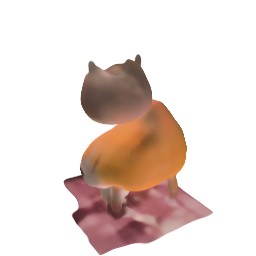} &
        \includegraphics[width=0.15\textwidth]{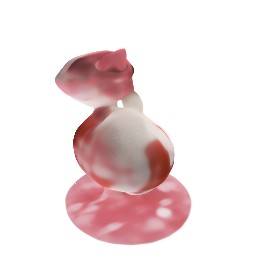} &
        \includegraphics[width=0.15\textwidth]{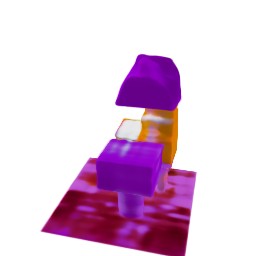} \\
        
        % \raisebox{0.05\textwidth}{$s=0.001$} &
        % \includegraphics[width=0.15\textwidth]{figures/guidance/a_corgi/0.001} &
        % \includegraphics[width=0.15\textwidth]{figures/guidance/a_corgi_wearing_a_santa_hat/0.001} &
        % \includegraphics[width=0.15\textwidth]{figures/guidance/a_red_cube_on_top_of_a_blue_cube/0.001} \\
        
        \raisebox{0.05\textwidth}{$s=0.01$} &
        \includegraphics[width=0.15\textwidth]{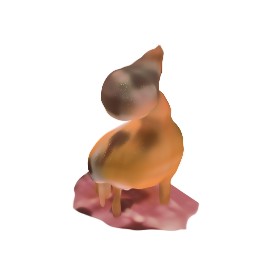} &
        \includegraphics[width=0.15\textwidth]{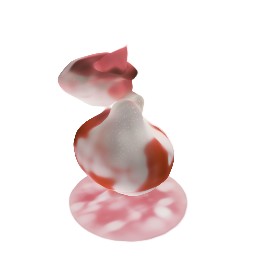} &
        \includegraphics[width=0.15\textwidth]{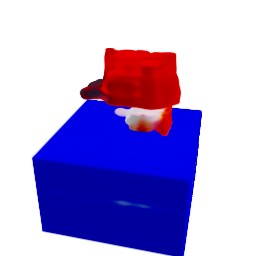} \\
        
        \raisebox{0.05\textwidth}{$s=0.05$} &
        \includegraphics[width=0.15\textwidth]{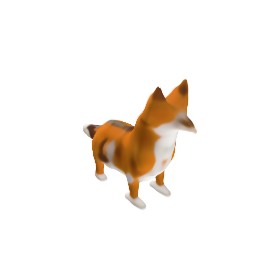} &
        \includegraphics[width=0.15\textwidth]{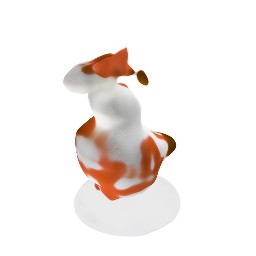} &
        \includegraphics[width=0.15\textwidth]{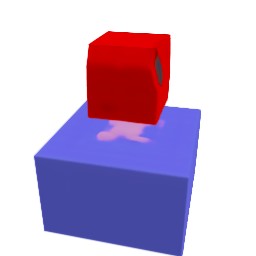} \\
        
        \raisebox{0.05\textwidth}{$s=0.1$} &
        \includegraphics[width=0.15\textwidth]{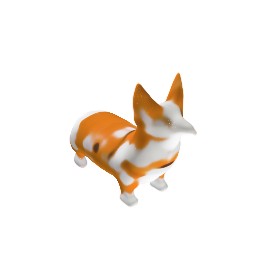} &
        \includegraphics[width=0.15\textwidth]{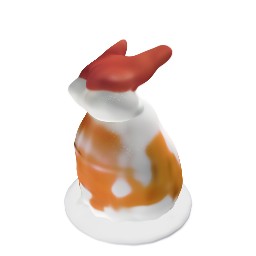} &
        \includegraphics[width=0.15\textwidth]{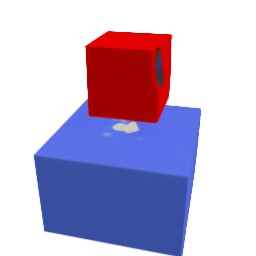} \\
        
        \raisebox{0.05\textwidth}{$s=0.5$} &
        \includegraphics[width=0.15\textwidth]{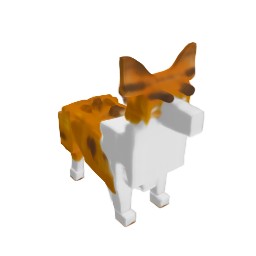} &
        \includegraphics[width=0.15\textwidth]{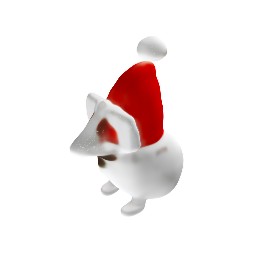} &
        \includegraphics[width=0.15\textwidth]{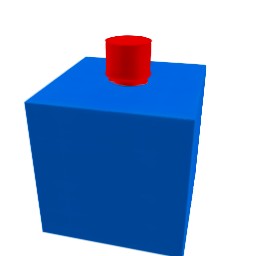} \\
        
        \raisebox{0.05\textwidth}{$s=1.0$} &
        \includegraphics[width=0.15\textwidth]{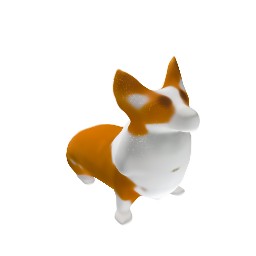} &
        \includegraphics[width=0.15\textwidth]{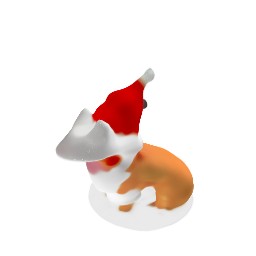} &
        \includegraphics[width=0.15\textwidth]{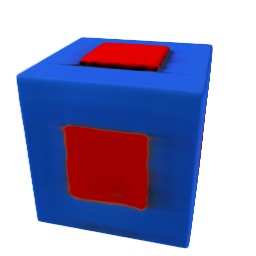} \\
    \end{tabular}
    \caption{\label{fig:guideddreamfusion} Using DreamFusion to guide our text-conditional \modelname{} model.}
\end{figure}

While our diffusion models operate in a latent space, we find that it is possible to guide them directly in \textit{image space}. During sampling, we have some noised latent $x_t$ and a corresponding model prediction $x_0 = f(x_t)$. If we treat the model prediction as a latent vector and render it with NeRF to get an image $I$, we can compute gradients of any image-based objective function $L$ like so:

$$\frac{\partial L}{\partial x_t} = \frac{\partial L}{\partial I} \frac{\partial I}{\partial x_0} \frac{\partial x_0}{\partial x_t}$$

Given this gradient, we can then follow the classifier guidance setup of \namecite{sotapaper} to update each diffusion step in the direction of a scaled gradient $s \cdot \frac{\partial L}{\partial x_t}$.

To test this idea, we leverage DreamFusion \shortcite{dreamfusion} to obtain image-space gradients that incentivize rendered images to match a text prompt. Since DreamFusion requires a powerful text-to-image diffusion model, we use the 3 billion parameter GLIDE model \shortcite{glide}. We sample from our text-conditional \modelname{} model using 1,024 stochastic DDPM steps. At each step, we use eight rendered views of the NeRF to obtain an estimate of the DreamFusion gradient. We then scale this gradient by a hyperparameter $s$ before applying a guided sampling step. This process takes roughly 15 minutes on eight NVIDIA V100 GPUs.

In Figure \ref{fig:guideddreamfusion}, we explore what happens as we increase the DreamFusion guidance scale $s$ while keeping the diffusion noise fixed. We observe in general that this text-conditional \modelname{} model is not very good on its own with DDPM sampling, failing to capture the text prompts with $s=0$. However, as we increase $s$, we find that the samples tend to approach something more closely matching the prompt. Notably, this is despite the fact that we do not use most of the tricks employed by DreamFusion, such as normals-based shading or grayscale rendering.

\end{document}